\documentclass[10pt,twocolumn,letterpaper]{article}
\usepackage{iccv}
\usepackage{times}
\usepackage{graphicx}
\usepackage{amssymb}
\usepackage{amsmath}
\usepackage{booktabs}
\usepackage{multicol}
\usepackage{multirow}
\usepackage{amsfonts}       
\usepackage{nicefrac}       
\usepackage{microtype}      
\usepackage{xspace}
\usepackage{graphicx}
\usepackage{epsfig}
\usepackage{algorithm}
\usepackage{color, colortbl}
\usepackage{algorithmic}
\usepackage{tablefootnote,footnotehyper}
\usepackage{array}
\usepackage[dvipsnames]{xcolor}
\usepackage{cases}
\newcolumntype{L}[1]{>{\raggedright\arraybackslash}p{#1}}
\newcolumntype{C}[1]{>{\centering\arraybackslash}p{#1}}
\newcolumntype{R}[1]{>{\raggedleft\arraybackslash}p{#1}}

\usepackage[pagebackref=true,breaklinks=true,letterpaper=true,colorlinks,bookmarks=false]{hyperref}

\iccvfinalcopy 

\def\iccvPaperID{6096} 

\ificcvfinal\fi

\begin{document}

\title{Domain Generalization via Rationale Invariance}

\author{Liang Chen$^1$\quad 
Yong Zhang$^{2}$\thanks{Corresponding  authors. This work is done when L. Chen is an intern in Tencent AI Lab.}\quad 
Yibing Song$^3$\quad 
Anton van den Hengel$^{1}$\quad 
Lingqiao Liu$^{1\ast}$\\
 {$^1$~The University of Adelaide}\quad  {$^2$~Tencent AI Lab}\quad
 {$^3$~AI$^3$ Institute, Fudan University}\\
 {\tt\small \{liangchen527, zhangyong201303, yibingsong.cv\}@gmail.com} \\ ~~~~~{\tt\small \{anton.vandenhenge, lingqiao.liu\}@adelaide.edu.au}
}

\maketitle
\ificcvfinal\thispagestyle{empty}\fi

\newcommand{\red}[0]{\textcolor{red}}
\newcommand{\blue}[0]{\textcolor{blue}}
\newcommand{\gray}[0]{\textcolor{gray}}
\newcommand{\rot}[0]{\rotatebox{90}}
\begin{abstract}
   This paper offers a new perspective to ease the challenge of domain generalization, which involves maintaining robust results even in unseen environments. Our design focuses on the decision-making process in the final classifier layer. Specifically, we propose treating the element-wise contributions to the final results as the rationale for making a decision and representing the rationale for each sample as a matrix. For a well-generalized model, we suggest the rationale matrices for samples belonging to the same category should be similar, indicating the model relies on domain-invariant clues to make decisions, thereby ensuring robust results. To implement this idea, we introduce a rationale invariance loss as a simple regularization technique, requiring only a few lines of code. Our experiments demonstrate that the proposed approach achieves competitive results across various datasets, despite its simplicity. Code is available at \url{https://github.com/liangchen527/RIDG}.
\end{abstract}

\section{Introduction}

Most existing machine learning models implicitly assume the training and test data are drawn from a similar distribution. While in practice, the real-world test samples often exhibit different characteristics due to distribution shift~\cite{koh2021wilds,ye2022ood}, resulting in an unsatisfactory performance for the deployed model. This limitation hinders the further application of deep models in various tasks, such as autonomous driving or object recognition. Hence, it is crucial importance to develop effective domain generalization (DG) methods that can maintain robust results regardless of domain shift.


\begin{figure}
    \centering
    \begin{minipage}[b]{0.32\linewidth}
		\centering
		\centerline{
			\includegraphics[width =\linewidth, height=1.5cm]{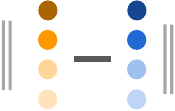}}
		\centerline{(a) Feature invariance}	
	\end{minipage} \hspace{0.5 cm}
	\begin{minipage}[b]{0.48\linewidth}
		\centering
		\centerline{
			\includegraphics[width =\linewidth, height=1.5cm]{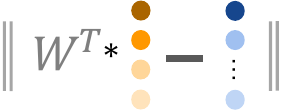}}
		\centerline{(b) Logit invariance}	
	\end{minipage} \\ \vspace{0.3cm}
	\begin{minipage}[b]{0.8\linewidth}
		\centering
		\centerline{
			\includegraphics[width =\linewidth, height=1.5cm]{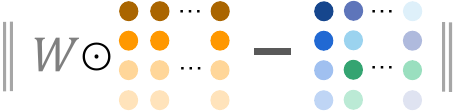}}
		\centerline{(c) Rational invariance (ours)}	
	\end{minipage}
 \caption{Visualized versions of different invariance regularizations. Here $\Vert \cdot \Vert$ are the $l_2$ norm; left and right sides of figures (a) - (c) denote the different feature, logit, and rational elements from the sample and the corresponding mean value; $\mathbf{W}$ is the weight in the classifier; $\odot$ is the element-wise product. In our setting, the rational matrix contains all element-wise contributions to the final results. Different from feature and logit invariance regularization, our rational invariance term considers both the feature and classifier weights involved in making a decision, providing a fine-grained characterization of the decision-making process.}
    \label{fig 1}
    \vspace{-0.4cm}
\end{figure}

The seminal work\cite{ben2006analysis} theoretically proves that well-generalized representations should remain consistent across different environments. Following this principle, various DG methods have been proposed to identify invariant features that are stable across diverse domains. These efforts include explicit feature alignment\cite{muandet2013domain, ghifary2016scatter, li2018domain,hu2020domain, atzmon2020causal}, domain-adversarial training~\cite{ganin2016domain,li2018domain,yang2021adversarial,li2018deep}, gradient regularization~\cite{arjovsky2019invariant,koyama2020invariance,shi2021gradient,rame2021ishr}, and meta-learning skills~\cite{li2018learning,balaji2018metareg,dou2019domain,li2019episodic}, to name a few. Despite some notable achievements, however, DG remains a formidable challenge and is far from being solved. In fact, a recent study \cite{gulrajani2020search} reveals that most current state-of-the-art methods perform inferior to the baseline empirical risk minimization (ERM) method\cite{vapnik1999nature} when applied with controlled model selection and restricted hyperparameter search. This finding highlights the need for innovative and effective models capable of maintaining robustness.

This work takes a different path toward achieving robust outputs by emphasizing the decision-making process in the classifier layer of a deep neural network, rather than focusing solely on the features. For most existing models, the final output logits are computed by multiplying the penultimate layer's feature with the classifier's weights\footnote{For simplicity, we omit the bias in the classifier.}. Delving deep into the process, each logit value can be regarded as the summation of the element-wise products between the feature and the corresponding weight. Considering each product term as a contribution to the corresponding logit, we collect all these contributions for  logits from all classes as a matrix and then reinterpret them as the rationale for making decisions regarding the input sample.

By introducing the new concept of rationale, we can refine and extend the invariance principle~\cite{ben2006analysis} from a new perspective. We posit that to ensure robust results, a well-generalized model should make decisions based on clues that are stable across samples and domains. Building on this intuition, we propose a regularization term that enforces similarity between the rationale matrix from each sample and the mean rationale matrix for the corresponding class. To implement our idea, we dynamically calculate the class-wise mean rationale matrix through momentum updates during training, which can be implemented with just a few lines of code.
As depicted in Figure~\ref{fig 1}, our approach differs from previous feature-based regularization~\cite{cha2022domain} in that we also consider the influence of the classifier, preventing biased estimation of feature importance. Additionally, by providing a more fine-grained characterization of the decision-making process, our model overcomes the limitation of logit-based regularization~\cite{pezeshki2021gradient}, which fails to account for the varying impacts of each contribution to the final decision.
Our experimental study also demonstrates that the proposed rationale invariance regularization strategy outperforms the feature and the logits invariance regularization schemes (see Figure~\ref{fig tease}).

By conducting extensive experiments on both the DomainBed~\cite{gulrajani2020search} and Wilds~\cite{koh2021wilds} benchmarks, we show that the proposed method consistently improves upon the baseline method and achieves comparable performance against state-of-the-art models. These results highlight the effectiveness of our new idea, despite its simplicity.

The contributions of this work are three-fold:
\begin{itemize}
    \item We introduce the concept of rationale in the decision-making process, which is new in the literature to the best of our knowledge, to improve DG.
    \item We propose a simple-but-effective strategy for utilizing the rationale concept toward the robust output objective, which is conducted by enforcing consistency between the rationale matrix from each sample and its corresponding mean value. 
    \item We conduct extensive experiments on the existing benchmark with rigorous evaluation protocol~\cite{gulrajani2020search} and demonstrate that the proposed rationale-based method can perform favorably against existing arts.
\end{itemize}

\def\swone{0.97\linewidth}
\begin{figure}
    \centering
    \begin{tabular}{c}
    \centering
    \includegraphics[width=\swone]{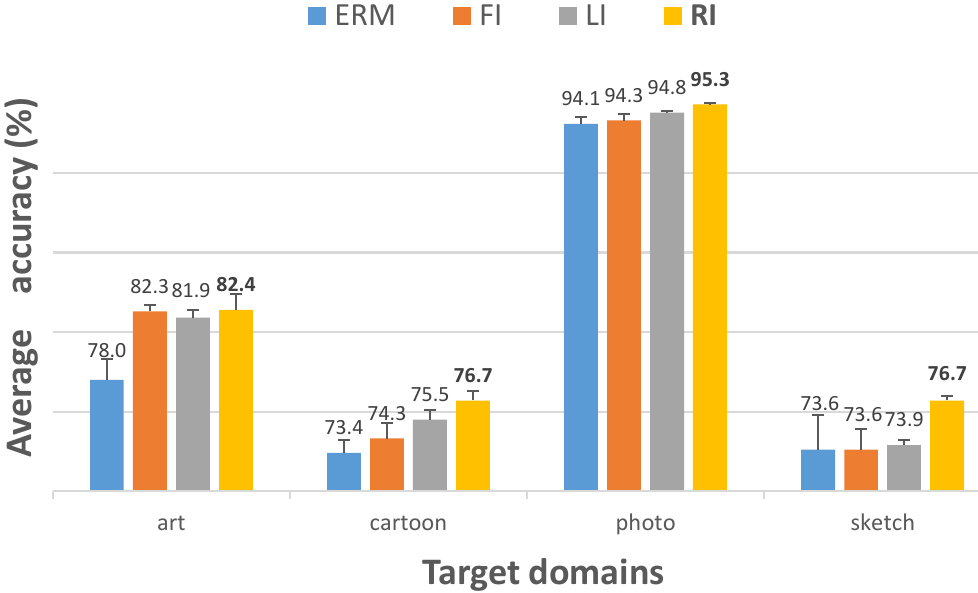}
    \end{tabular}
    \vspace{-0.1 cm}
    \caption{Performance improvements from different invariance regularization, \ie feature invariance (FI), logit invariance (LI), and the proposed rationale invariance (\textbf{RI}), for the baseline ERM model (\ie implemented with ResNet18 \cite{he2016deep} backbone). Experiments are conducted on the PACS dataset \cite{li2017deeper} with the leave-one-out setting. Following \cite{gulrajani2020search}, we use 60 sets of random seeds and hyper-parameters for each target domain. The reported average accuracy and error bars verify the effectiveness of our method.}
    \label{fig tease}
    \vspace{-0.3 cm}
\end{figure}

\section{Related Works}
Domain generalization (DG), designed to enable a learned model to maintain robust results in unseen target domains, is gaining increasing attention in the research community lately. The problem can be traced back to a decade ago \cite{blanchard2011generalizing}, and various approaches have been proposed to push the generalization boundary ever since~\cite{muandet2013domain,ghifary2016scatter, li2018domain,hu2020domain,ganin2016domain,li2018domain,yang2021adversarial,li2018deep,li2018learning,balaji2018metareg,dou2019domain,li2019episodic,rame2021ishr,pezeshki2021gradient,chen2022self,chen2022comen,chen2022ost,chen2022mix,chen2023activate,chen2023improved}. The primary goal for most current arts is to identify invariant features given the limited training domains~\cite{atzmon2020causal,kim2021selfreg}. 

Despite the varying details, we can categorize them into the following types.
\noindent\textbf{(1) Explicit mining invariant features.} The previous work~\cite{ben2006analysis} theoretically reveals that if the features remain invariant across different environments, then they are general and transferable to different domains. To this end, some existing arts aim to explicitly explore invariant features for generalization. For example, \cite{muandet2013domain} suggests using maximum mean discrepancy (MMD) to align the obtained features from different domains; \cite{ghifary2015domain} adopts a multi-task auto-encoder that transforms the original image into analogs in multiple related domains and expects the learned features to be robust across domains. The idea of explicitly ensuring invariance has been explored in the gradient level recently, such as that in \cite{shi2021gradient} and \cite{rame2021ishr}, where the inner products of gradients from different domains are maximized to achieve robustness \cite{shi2021gradient}, and gradients of samples from different domains are expected to be similar to their mean values \cite{rame2021ishr}.  
\textbf{(2) Using specially-designed optimization algorithms.} There are also approaches that resort to different training strategies, such as adversarial learning \cite{ganin2016domain,li2018domain,yang2021adversarial,li2018deep} and meta-learning \cite{li2018learning,balaji2018metareg,dou2019domain,li2019episodic}, for finding invariance features. For example, adversarial training is adopted in \cite{ganin2016domain} to enforce the learned features to be agnostic about the domain information, where the target mainstream feature that minimizes the semantic classification loss is also required to maximize the domain classification loss. This idea is combined with the MMD constraint in \cite{li2018deep} to update an auto-encoder. Based on the MAML framework \cite{finn2017model}, most of the meta-learning-based techniques try to find invariance by simulating distribution shifts between seen and unseen environments \cite{li2018learning,balaji2018metareg,dou2019domain,li2019episodic}. In \cite{huang2020self}, Huang \etal suggest learning diverse features via masking out the easy clues during training. Test-time training is also introduced in \cite{chen2023improved} for the task of DG.
\textbf{(3) Augmentation.} Existing augmentation techniques applied in DG are often focused on two folds, the feature level \cite{li2021simple,zhou2021domain,nam2021reducing} that combines different features or their style statistics, and image level \cite{yan2020improve,xu2021fourier} which synthesize new data by directly mixing images \cite{yan2020improve} or its phase \cite{xu2021fourier}. It is noteworthy that not all augmentation skills are effective, only those related to the data from test domains can benefit generalization~\cite{shorten2019survey}, explaining why these methods can not ensure improvements for DG. 
 
However, despite a proliferation of DG models, doubts are cast regarding the improvements of current arts compared with the baseline model~\cite{gulrajani2020search}. Indeed, both the results in \cite{gulrajani2020search} and our experimental studies suggest most current  methods cannot improve DG in all evaluated datasets. In contrast, our method is consistently beneficial for the baseline method. Moreover, compared to the related methods that use feature~\cite{cha2022domain} or logit~\cite{pezeshki2021gradient} invariance regularizations, our method also shows certain improvements, demonstrating the effectiveness of the new rationale concept.

\section{Methdology}

\noindent\textbf{Preliminary.} Let $\mathcal{D}_s=\{\mathcal{D}_1, \mathcal{D}_2,...,\mathcal{D}_M\}$ be a set of given training domains, where $\mathcal{D}_i$ is a joint distribution over the image space $\mathcal{X}$ and label space $\mathcal{Y}$. From each domain $\mathcal{D}_i$, we can observe $n_i$ training data points which consist of input $x \in \mathcal{X}$ and semantic label $y \in \mathcal{Y}$: $(x_j^i, y_j^i)_{j=1}^{n_i} \sim \mathcal{D}_i$. Given the target domain $\mathcal{D}_t$ that samples images from a different distribution, the vanilla DG task asks a model trained on the multi-source $\mathcal{D}_s$ (with $M > 1$) to perform well on the unseen domain $\mathcal{D}_t$. 

\subsection{Rationale in the Decision-Making Process}
This section includes a comprehensive description of our newly proposed rationale representation, as well as comparisons against existing concepts.

In current deep models, the final output results derived from the decision-making process involve obtaining the feature $\mathbf{z} \in \mathbb{R}^{D}$ through a feature extractor $f$ (\ie $\mathbf{z} = f(x)$), followed by the computation of the final logits $\mathbf{o}\in \mathbb{R}^K$ by applying the classifier $h$ on $\mathbf{z}$ (\ie $\mathbf{o} = h(\mathbf{z})$).
Assuming there are $K$ classes in $\mathcal{Y}$, the output logits can be represented as $\mathbf{o} = \mathbf{W}^T \mathbf{z} \in \mathbb{R}^K$, where $\mathbf{W} \in \mathbb{R}^{D\times K}$ is the weight of $h$. To delve deeper into this process, we can represent each logit value (the $k$-th dimension of $\mathbf{o}$) $o_k$ as generated by the summation of element-wise products between the feature elements and the corresponding weights in the classifier layer, which can be represented as following:
\begin{equation}
    \label{eq evidence}
    o_k = \mathbf{W}^\top_{\{, k\}} \mathbf{z} = \sum_{j=1}^D W_{\{j, k\}} z_j.
\end{equation}
In the above equation, the logit for class $k$ is simply the accumulation of all $W_{\{j, k\}} z_j$ which can be regarded as an element-wise contribution to $o_k$. Intuitively, $W_{\{j, k\}} z_j$ provides a piece of evidence to support the specific decision. We thus can collect all $W_{\{j, k\}} z_j$ as a matrix $\mathbf{R} \in \mathbb{R}^{D \times K}$ to represent the rationale of classifying a sample, which is formally defined as following:
\begin{equation}
\label{eq rationale}
\mathbf{R} = 
\begin{bmatrix}
W_{\{1, 1\}} z_1 &W_{\{1, 2\}} z_1  &$\dots$ &W_{\{1, K\}} z_1 \\
W_{\{2, 1\}} z_2 &W_{\{2, 2\}} z_2 &$\dots$ &W_{\{2, K\}} z_2 \\
\vdots &\vdots & \ddots &\vdots \\
W_{\{D, 1\}} z_D &W_{\{D, 2\}} z_D  &$\dots$ &W_{\{D, K\}} z_D \\
\end{bmatrix}.
\end{equation}
Note that we construct the rational matrix $\mathbf{R}$ from all classes. This is because the final posterior probability is often calculated via performing the Softmax function on all the logits. In other words, logits from all classes will affect the final posterior probability.

\noindent\textbf{Rationale V.S. feature V.S. logit:}
To ensure robust results, previous studies explicitly or implicitly regularize the decision-making process through the analysis of features~\cite{cha2022domain} or logits~\cite{pezeshki2021gradient}. However, these two strategies have intrinsic limitations. Specifically, by focusing only on features, the classifier weights that determine the effects of different feature elements may not be given sufficient consideration, leading to biased estimates of feature importance. For instance, a feature element with a large value may correspond to a small value in the classifier, resulting in a lower impact on the final results. Simply looking at the feature value or variance can therefore be misleading, thus diminishing the overall performance of the model.

In contrast to features, logit implicitly takes the classifier into account, alleviating the issue in the situation that focuses only on features to a certain extent. However, the logit provides only a coarse decision value and lacks a detailed explanation of the decision-making process. This limitation makes it challenging to comprehend the rationale behind a decision. Consequently, solely relying on logits may not provide control over the decision-making process, leading to ineffectiveness in obtaining robust outputs. This limitation is further validated in our visual analysis in Sec.~\ref{sec tsne}.

Our rationale matrix representation, i.e., $R$ matrix, overcomes the limitations of both feature-based and logit-based analyses by representing the decision-making process from a more fine-grained perspective that also takes the classifier into account. It provides a new tool for domain generalization or other related machine learning problems. We provide more analysis regarding these strategies in Sec.~\ref{sec ana_rational}.

\newcommand{\D}{\hspace{0.4 cm}}
\newcommand{\comment}{\textcolor{TealBlue}}
\begin{algorithm}[t]
\caption{Pseudo codes of the training process for the proposed method in a PyTorch-like style.}
\footnotesize
\comment{\# ~~ $f, h$: ~feature extractor, classifier}\\
\comment{\# ~~ $K, D$: ~total number of classes, dimension of the feature}\\
\comment{\# ~~ $m, \alpha$: ~momentum, weight parameter} \vspace{0.3 cm}\\
$\overline{R}$ = torch.zeros($K$, $K$, $D$)~~\comment{\# ~~initialize $\overline{R}$}\\
\comment{\# ~~training process}\\
for $x, y$ in training\_loader: ~~\comment{\# ~~load a minibatch with $N$ samples}\\
\begin{algorithmic}[]
\vspace{-0.4 cm}
\STATE $z$ = $f(x)$ \vspace{0.3CM}\\
\comment{\# ~~obtain the rationale tensor for the current batch}\\
\STATE $R$ = torch.zeros($K$, $N$, $D$) ~~\comment{\# ~~initialize $R$}\\
\comment{\# ~~obtaining $R$ based on Eq.~\eqref{eq rationale}}\\
\STATE for $i$ in range(K):
\STATE \D $R\left[i\right]$ = $h$.weight$\left[i\right] \ast z$ \vspace{0.3 cm}\\
\STATE classes = torch.unique($y$) \\
\STATE $\mathcal{L}_{inv}=0$ \\
\comment{\# ~~compute $\mathcal{L}_{inv}$ for different classes}\\
\STATE for $i$ in range(classes.shape[0]):
\STATE \D \comment{\# ~~mean values of $R$ in the current batch}\\
\STATE \D $R_{\text{mean}} = R\left[:, y==\text{classes}\left[i\right]\right]$.mean(dim=1).detach()\\
\STATE \D \comment{\# ~~update $\overline{R}$ for different classes based on Eq.~\eqref{eq momentum}}\\
\STATE \D $\overline{R}\left[\text{classes}\left[i\right] \right]$ = $(1 - m) \ast \overline{R}\left[\text{classes}\left[i\right] \right]$ + $m \ast R_{\text{mean}}$ \vspace{0.3 cm}\\
\STATE \D \comment{\# ~~computing $\mathcal{L}_{inv}$ based on Eq.~\eqref{eq inv}}\\
\STATE \D $\mathcal{L}_{inv} += \text{MSELoss}(R\left[:, y==\text{classes}\left[i\right]\right],~~\overline{R}\left[\text{classes}\left[i\right] \right])$ \vspace{0.3 cm}\\

\STATE \comment{\# ~~computing $\mathcal{L}_{all}$ based on Eq.~\eqref{eq all}}\\
\STATE $\mathcal{L}_{all}$ = CrossEntropyLoss($h(z), y$) + $\alpha \mathcal{L}_{inv}$\\
\STATE ($\left[f.\text{params},~~ h.\text{params}\right]$).zero\_grad()\\
\STATE $\mathcal{L}_{all}$.backward()\\
\STATE update($\left[f.\text{params},~~ h.\text{params}\right]$)
\end{algorithmic}
\label{alg 1}
\end{algorithm}

\subsection{Rationale Invariance for DG}
Our primary goal is to design a model that can maintain robust outputs despite the varying environments.
The rationale matrix provides a characterization of how a decision is made, and new regularization can be introduced upon the rationale concept to control the decision-making process to approximate the optimal design. 
Various methods can be potentially used for designing such regularization terms, but this study adopts a simple approach by ensuring that the rationale matrix of samples from the same class aligns with their corresponding mean value. This approach implies that the decision to classify an object should be based on the same reasoning, regardless of the variation in samples or domains. This is reasonable because the causal factors for decision-making are often stable patterns that persist across samples and domains~\cite{mahajan2021domain}.

Formally, we impose an invariance loss term to enforce the rationale matrix from a sample to be close to its corresponding mean. Denoting as $\mathcal{L}_{inv}$, this regularization term can be written as:
\begin{equation}
    \label{eq inv}
    \mathcal{L}_{inv} = \frac{1}{N_b}\sum_k \sum_{\{n|y_n=k\}} \Vert \mathbf{R}_n - \overline{\mathbf{R}}_k \Vert^2,
\end{equation}
where $\Vert \cdot \Vert$ is the $l2$ norm, $N_b$ is the number of samples in a mini-batch; $\mathbf{R}_n$ denotes the rationale matrix for the $n$-th sample.  $\overline{\mathbf{R}}_k$ denotes the mean matrix of the rationale matrix corresponding to the $k$-th class.

However, calculating $\overline{\mathbf{R}}_k$ directly involves averaging through all the samples, which is computationally expensive. Thus, we suggest updating $\overline{\mathbf{R}}_k$ online in each iteration step in a momentum fashion. Inspired by the previous work~\cite{he2020momentum}, the updating process for $\overline{\mathbf{R}}_k$ can be given as:
\begin{equation}
    \label{eq momentum}
    \overline{\mathbf{R}}_k^t = (1 - m)\times \overline{\mathbf{R}}_k^{t-1} + m \times \frac{1}{\vert y_n=k \vert}\sum_{\{n|y_n=k\}} \mathbf{R}_n,
\end{equation}
where $t$ is the iteration index, $m$ is a positive momentum value, and $\vert y_n=k \vert$ computes the sample corresponding to the $k$-th class. We initialize $\overline{\mathbf{R}}_k$ with the rationale computed from the first iteration step.

\subsection{Learning Objective}
Combining the rationale invariance constraint with the classification loss $\mathcal{L}_{cla} = \frac{1}{N_b} \sum_n \text{CE}(h(f(x_n)), y_n)$, where CE denotes the cross-entropy loss, our overall training objective $\mathcal{L}_{all}$ merely contains two terms,
\begin{equation}
    \label{eq all}
    \mathcal{L}_{all} = \mathcal{L}_{cla} + \alpha \mathcal{L}_{inv},
\end{equation}
where $\alpha$ is a positive weight. The optimization task regarding minimization Eq.~\eqref{eq all} can be fulfilled by the existing technique~\cite{kingma2014adam}. The pseudo-code for the training process is presented in Algorithm~\ref{alg 1}. As seen, the proposed method is extremely simple, as it only adds a few lines based on the standard ERM training pipeline. 

\section{Experiments}

\begin{table*}[t]
\centering
\caption{Evaluations on the DomainBed benchmark \cite{gulrajani2020search}. All methods are examined for 60 trials in each unseen domain. Here Top5 accumulates the number of datasets where a method achieves the top 5 performances. Every symbol $\uparrow$ denotes a score of $+1$, meaning the specific method outperforms ERM (on account of their variances), and vice versa for the symbol $\downarrow$, which denotes a score of $-1$; otherwise, the score is 0. The best results are colored as \red{red}, and the second bests are colored as \blue{blue}}. 
\scalebox{1}{
\begin{tabular}{lC{1.8cm}C{1.8cm}C{1.8cm}C{1.8cm}C{1.8cm}|C{0.6cm} C{0.7cm} C{0.7cm}}
\toprule 
& PACS & VLCS & OfficeHome & TerraInc & DomainNet & Avg. & Top5 &Score\\
\hline \hline
MMD \cite{li2018domain} &81.3 $\pm$ 0.8$\uparrow$ &74.9 $\pm$ 0.5$\downarrow$ &59.9 $\pm$ 0.4$\downarrow$ &42.0 $\pm$ 1.0$\uparrow$ &7.9 $\pm$ 6.2$\downarrow$ &53.2 &1 &-1\\
RSC  \cite{huang2020self} &80.5 $\pm$ 0.2$\uparrow$ &75.4 $\pm$ 0.3 &58.4 $\pm$ 0.6$\downarrow$ &39.4 $\pm$ 1.3 &27.9 $\pm$ 2.0$\downarrow$ &56.3 &0 &-1\\
IRM \cite{arjovsky2019invariant} &80.9 $\pm$ 0.5$\uparrow$ & 75.1 $\pm$ 0.1$\downarrow$ &58.0 $\pm$ 0.1$\downarrow$ &38.4 $\pm$ 0.9 &30.4 $\pm$ 1.0$\downarrow$ &56.6 &0 &-2\\
ARM \cite{zhang2020adaptive} &80.6 $\pm$ 0.5 &75.9 $\pm$ 0.3 &59.6 $\pm$ 0.3$\downarrow$ &37.4 $\pm$ 1.9 &29.9 $\pm$ 0.1$\downarrow$ &56.7 &0 &-2\\
DANN \cite{ganin2016domain} &79.2 $\pm$ 0.3 &76.3 $\pm$ 0.2$\uparrow$ &59.5 $\pm$ 0.5$\downarrow$ &37.9 $\pm$ 0.9 &31.5 $\pm$ 0.1$\downarrow$ &56.9 &1 &-1\\
GroupGRO \cite{sagawa2019distributionally} &80.7 $\pm$ 0.4$\uparrow$ &75.4 $\pm$ 1.0 &60.6 $\pm$ 0.3 &41.5 $\pm$ 2.0 &27.5 $\pm$ 0.1$\downarrow$ &57.1 &0 &-1\\
CDANN \cite{li2018deep} &80.3 $\pm$ 0.5 &76.0 $\pm$ 0.5 &59.3 $\pm$ 0.4$\downarrow$ &38.6 $\pm$ 2.3 &31.8 $\pm$ 0.2$\downarrow$ &57.2 &0 &-2\\
VREx \cite{krueger2021out} &80.2 $\pm$ 0.5 &75.3 $\pm$ 0.6 &59.5 $\pm$ 0.1$\downarrow$ &\blue{43.2 $\pm$ 0.3}$\uparrow$ &28.1 $\pm$ 1.0$\downarrow$ &57.3 &1 &-1\\
CAD \cite{ruan2021optimal} &81.9 $\pm$ 0.3$\uparrow$ &75.2 $\pm$ 0.6 &60.5 $\pm$ 0.3 &40.5 $\pm$ 0.4$\uparrow$ &31.0 $\pm$ 0.8$\downarrow$ &57.8 &1 &1\\
CondCAD \cite{ruan2021optimal} &80.8 $\pm$ 0.5$\uparrow$ &76.1 $\pm$ 0.3 &61.0 $\pm$ 0.4 &39.7 $\pm$ 0.4 &31.9 $\pm$ 0.7$\downarrow$ &57.9 &0 &0\\
MTL \cite{blanchard2017domain} &80.1 $\pm$ 0.8 &75.2 $\pm$ 0.3$\downarrow$ &59.9 $\pm$ 0.5 &40.4 $\pm$ 1.0 &35.0 $\pm$ 0.0$\downarrow$ &58.1 &0 &-2\\
ERM \cite{vapnik1999nature} &79.8 $\pm$ 0.4 &75.8 $\pm$ 0.2 &60.6 $\pm$ 0.2 &38.8 $\pm$ 1.0 &35.3 $\pm$ 0.1 &58.1 &0 &-\\
MixStyle \cite{zhou2021domain} &\blue{82.6 $\pm$ 0.4}$\uparrow$ &75.2 $\pm$ 0.7 &59.6 $\pm$ 0.8 &40.9 $\pm$ 1.1 &33.9 $\pm$ 0.1$\downarrow$ &58.4 &1 &0\\
MLDG \cite{li2018learning} &81.3 $\pm$ 0.2$\uparrow$ &75.2 $\pm$ 0.3$\downarrow$ &60.9 $\pm$ 0.2 &40.1 $\pm$ 0.9 &35.4 $\pm$ 0.0 &58.6 &0 &0\\
Mixup \cite{yan2020improve} &79.2 $\pm$ 0.9 &76.2 $\pm$ 0.3 &61.7 $\pm$ 0.5 &42.1 $\pm$ 0.7$\uparrow$ &34.0 $\pm$ 0.0$\downarrow$ &58.6 &1 &0\\
MIRO \cite{cha2022domain} &75.9 $\pm$ 1.4$\downarrow$ &\blue{76.4 $\pm$ 0.4} &\red{64.1 $\pm$ 0.4}$\uparrow$ &41.3 $\pm$ 0.2$\uparrow$ &\blue{36.1 $\pm$ 0.1}$\uparrow$ &58.8 &3 &2 \\
Fishr  \cite{rame2021ishr} &81.3 $\pm$ 0.3$\uparrow$ &76.2 $\pm$ 0.3 &60.9 $\pm$ 0.3 &42.6 $\pm$ 1.0$\uparrow$ &34.2 $\pm$ 0.3$\downarrow$ &59.0 &1 &1\\
SagNet \cite{nam2021reducing} &81.7 $\pm$ 0.6$\uparrow$ &75.4 $\pm$ 0.8 &62.5 $\pm$ 0.3$\uparrow$ &40.6 $\pm$ 1.5 &35.3 $\pm$ 0.1 &59.1 &1 &2\\
SelfReg \cite{kim2021selfreg} &81.8 $\pm$ 0.3$\uparrow$ &\blue{76.4 $\pm$ 0.7} &62.4 $\pm$ 0.1$\uparrow$ &41.3 $\pm$ 0.3$\uparrow$ &34.7 $\pm$ 0.2$\downarrow$ &59.3 &2 &2\\
Fish \cite{shi2021gradient} &82.0 $\pm$ 0.3$\uparrow$ &\red{76.9 $\pm$ 0.2}$\uparrow$ &62.0 $\pm$ 0.6$\uparrow$ &40.2 $\pm$ 0.6 &35.5 $\pm$ 0.0$\uparrow$ &59.3 &3 &4\\
CORAL \cite{sun2016deep} &81.7 $\pm$ 0.0$\uparrow$ &75.5 $\pm$ 0.4 &62.4 $\pm$ 0.4$\uparrow$ &41.4 $\pm$ 1.8 &\blue{36.1 $\pm$ 0.2}$\uparrow$ &59.4 &2 &3\\
SD \cite{pezeshki2021gradient} &81.9 $\pm$ 0.3$\uparrow$ &75.5 $\pm$ 0.4 &{62.9 $\pm$ 0.2}$\uparrow$ &42.0 $\pm$ 1.0$\uparrow$ &\red{36.3 $\pm$ 0.2}$\uparrow$ &59.7 &4 &4\\
Ours &\red{82.8 $\pm$ 0.3}$\uparrow$ &75.9 $\pm$ 0.3 &\blue{63.3 $\pm$ 0.1}$\uparrow$ &\red{43.7 $\pm$ 0.5}$\uparrow$ &36.0 $\pm$ 0.2$\uparrow$ &60.3 &4 &4\\
\bottomrule
\end{tabular}}
\label{tab domainbed}
\vspace{-0.3 cm}
\end{table*}

\subsection{Experiments on DomainBed~\cite{gulrajani2020search}}
\label{sec set}
\noindent\textbf{Datasets.}
We use five different datasets in the DomainBed benchmark~\cite{gulrajani2020search} to comprehensively evaluate the proposed method, namely PACS \cite{li2017deeper}, VLCS \cite{fang2013unbiased}, OfficeHome \cite{venkateswara2017deep}, TerraInc \cite{beery2018recognition}, and DomainNet~\cite{peng2019moment}. \textbf{(1) PACS} has 9,991 images that can be categorized into 7 classes. This dataset is probably the most commonly used dataset in the DG literature due to its large distributional shift across 4 domains, including art painting, cartoon, photo, and sketch; \textbf{(2) VLCS} consists of 10,729 images collected from 5 different classes. These images are originally from  4 different datasets (\ie PASCAL VOC 2007 \cite{everingham2010pascal}, LabelMe \cite{russell2008labelme}, Caltech \cite{fei2004learning}, and Sun \cite{xiao2010sun}), and each dataset is considered a domain in the DG setting;
\textbf{(3) OfficeHome} is an object recognition dataset that contains 15,588 images from 65 classes, which can be divided into 4 domains including artistic, clipart, product, and real world; \textbf{(4) TerraInc} consists of 24,788 animal images captured in the wild from different locations, there are a total of 10 classes in it with the corresponding location viewed as the varying domain, \ie L100, L38, L43, L46.
\textbf{(5) DomainNet} contains 586,575 images from a total of 345 classes, whose domains can be depicted into 6 types: clipart, infograph, painting, quickdraw, real, and sketch.

\noindent\textbf{Implementation details.} Following the prevelant design, we use the imagenet~\cite{deng2009imagenet} pretrained ResNet18 model~\cite{he2016deep} as the backbone for all the datasets. Results with the ResNet50 backbone are presented in the supplementary material.
We use dynamic ranges for the value of momentum in Eq.~\eqref{eq momentum}: $m \in \left[0.0001, 0.1 \right]$ and the weight parameter in Eq.~\eqref{eq all}: $\alpha \in \left[0.001, 0.1 \right]$. Regarding other settings related to the compared arts: for all the datasets, we use the leave-one-out strategy to evaluate these methods, which uses one domain as the hold-out target domain and others as the source domains, and we evaluate all the compared methods each with 60 trials for different datasets. 
Specifically, in each trial, the training and test samples are randomly split in a ratio of 8:2 (trial:val), and all the compared methods are reevaluated using the default settings in DomainBed in the same device (8 Nvidia Tesla v100 GPUs and each with 32G memory) to ensure fair comparisons. The hyper-parameter settings, such as learning rate, augmentation strategies, and batch size, are all dynamically set according to~\cite{gulrajani2020search}.

\begin{table*}[t]
    \centering
    \caption{Evluations on four challenging datasets from the Wilds benchmark~\cite{koh2021wilds}. Results are directly cited from the public leaderboard. Metrics of means and standard deviations are reported across different trials (according to the default settings in \cite{koh2021wilds}). The best results are colored as \red{red}. We reevaluate ERM (\ie ERM (ours)) as a comparison. Our method performs favorably against existing arts, leading in half of the evaluated datasets, and obtains better results than the corresponding ERM in all situations.}
    \scalebox{1}{
    \begin{tabular}{lC{1.8cm}C{1.8cm}C{1.8cm}C{1.8cm}C{1.8cm}C{1.8cm}}
\toprule 
& \multicolumn{2}{c}{iWildCam}\hspace{0pt} & Camelyon17\hspace{0pt} & RxRX1 & \multicolumn{2}{c}{FMoW} \\
  \cmidrule(lr){2-3} \cmidrule(lr){4-4} \cmidrule(lr){5-5} \cmidrule(lr){6-7}
    Method &Avg. acc. &Macro F1 &Avg. acc. & Avg. acc. &Worst acc.  & Avg. acc. \\
    \hline \hline
    ERM \cite{vapnik1999nature} & 71.6 $\pm$ 2.5 &31.0 $\pm$ 1.3 &70.3 $\pm$ 6.4 &29.9 $\pm$ 0.4 &32.3 $\pm$ 1.25 &53.0 $\pm$ 0.55\\
    CORAL~\cite{sun2016deep} &\red{73.3 $\pm$ 4.3} &\red{32.8 $\pm$ 0.1} &59.5 $\pm$ 7.7 &28.4 $\pm$ 0.3 &31.7 $\pm$ 1.24 &50.5 $\pm$ 0.36\\
    GroupGRO \cite{sagawa2019distributionally} & 72.7 $\pm$ 2.1 &23.9 $\pm$ 2.0 &68.4 $\pm$ 7.3 &23.0 $\pm$ 0.3 &30.8 $\pm$ 0.81 &52.1 $\pm$ 0.50\\
    IRM~\cite{arjovsky2019invariant} & 59.8 $\pm$ 3.7 &15.1 $\pm$ 4.9 &64.2 $\pm$ 8.1 &8.2 $\pm$ 1.1 &30.0 $\pm$ 1.37 &50.8 $\pm$ 0.13\\
    ARM-BN \cite{zhang2020adaptive} & 70.3 $\pm$ 2.4 &23.7 $\pm$ 2.7 &87.2 $\pm$ 0.9 &\red{31.2 $\pm$ 0.1} &24.6 $\pm$ 0.04 &42.0 $\pm$ 0.21\\
    TTBNA~\cite{schneider2020improving} & 46.6 $\pm$ 0.9 &13.8 $\pm$ 0.6 &- &20.1 $\pm$ 0.2 &30.0 $\pm$ 0.23 &51.5 $\pm$ 0.25\\
    Fish~\cite{shi2021gradient} & 64.7 $\pm$ 2.6 &22.0 $\pm$ 1.8 &74.7 $\pm$ 7.1 &- &34.6 $\pm$ 0.18 &51.8 $\pm$ 0.32\\
    CGD~\cite{piratla2021focus} &- &- &69.4 $\pm$ 7.9 &- &32.0 $\pm$ 2.26 &50.6 $\pm$ 1.39\\
    LISA~\cite{yao2022improving} & - &- &77.1 $\pm$ 6.9 &31.9 $\pm$ 1.0 &35.5 $\pm$ 0.81 &52.8 $\pm$ 1.15\\
    \hline
    ERM (ours) & 70.5 $\pm$ 4.3 &30.1 $\pm$ 0.7 &80.1 $\pm$ 3.9 &29.9 $\pm$ 0.3 &33.0 $\pm$ 1.34 &53.1 $\pm$ 1.16\\
    Ours & 70.9 $\pm$ 2.8 &30.7 $\pm$ 1.0 &\red{90.6 $\pm$ 2.9} &30.0 $\pm$ 0.3 &\red{36.1 $\pm$ 1.48} &\red{55.9 $\pm$ 0.25}\\
    \bottomrule
    \end{tabular}}
    \label{tab wilds}
    \vspace{-0.5 cm}
\end{table*}

\noindent\textbf{Experimental results.}
Average results from the 60 trials of different compared methods in the DomainBed benchmark are listed in Table~\ref{tab domainbed}. We use average accuracy (\ie Avg.), leading performances (\ie Top5), and performance score~\cite{ye2022ood} (\ie Score) to evaluate the compared arts. We note that the simple ERM baseline obtains favorable performance against existing arts in the term of average accuracy. As a matter of fact, among all the compared arts, less than half can benefit ERM in general (with a Score $>$ 0), and only 5 methods(\ie SagNet \cite{nam2021reducing}, Fish \cite{shi2021gradient}, CORAL \cite{sun2016deep}, SD \cite{pezeshki2021gradient}, and Ours) do not decrease the performance of ERM in all datasets. These results comply with the observation in~\cite{gulrajani2020search} and show that most existing strategies cannot improve ERM when evaluated with rigorous settings.
In comparison, the proposed rationale invariance framework leads the baseline ERM by 2 percent and the second best model (\ie SD~\cite{pezeshki2021gradient}) by 0.6 percent in the term of average accuracy. Meanwhile, it also obtains the best results in 2 out of the 5 benchmarks (\ie PACS and TerraInc datasets) and 4 in the top 5, performing one of the best in the Score criteria, demonstrating its effectiveness compared with existing models.

Moreover, when compared with arts that adopt the feature-invariance~\cite{cha2022domain} and logit-invariance~\cite{pezeshki2021gradient} regularizations, our method also showcases comparable or even better results, especially in the PACS and TerraInc datasets, demonstrating the effectiveness of the new rationale concept. Note that the feature-invariance inspired method MIRO~\cite{cha2022domain} performs inferior to others when evaluated in the PACS dataset. This is because their approach specifically enforces similarity between intermediate features from the model and that from the pretrained backbone, which can be detrimental to the performance when there is a significant distribution shift between the target data, such as cartoon' or 'sketch' images from PACS, and samples used for pertaining. 
We provide more experimental studies to analyze the invariance constraints from MIRO~\cite{cha2022domain} and SD~\cite{pezeshki2021gradient} in Sec.~\ref{sec ana_rational}.
Results of average accuracy in each domain from different datasets are presented in the supplementary material. Please also refer to it for further details.

\subsection{Experiments on Wilds~\cite{koh2021wilds}}
\noindent\textbf{Datasets.}
The Wilds benchmark~\cite{koh2021wilds} contains multiple datasets that capture real-world distribution shifts across a diverse range of modalities. We conduct experiments on four challenging datasets from Wilds to further examine our method, including iWildCam~\cite{beery2021iwildcam}, Camelyon17~\cite{bandi2018detection}, RxRx1~\cite{taylor2019rxrx1}, FMoW~\cite{christie2018functional}. Specifically, \textbf{(1) iWildCam} contains 203,029 animal images from 182 different species, which are taken from a total of 324 camera traps in different locations (\ie domains); \textbf{(2) Camelyon17} has 45,000 images used for the binary tumor classification task, which are collected from 5 different hospitals (\ie domains); \textbf{(3) RxRx1} consists of 125,514 high-resolution fluorescence microscopy images of human cells under 1,108 genetic perturbations (\ie classes) in 51 experimental batches (\ie domains); \textbf{(4) FMoW} is a satellite dataset contains 118,886 samples used for land classification across different regions and years, and the total categories and domains are 62 and 80 (\ie years $\times$ regions), respectively.   

\noindent\textbf{Implementation and evaluation details.}
Implementation details for different datasets vary in the Wilds benchmark~\cite{koh2021wilds}. We implement our model and the baseline ERM~\cite{vapnik1999nature} with the default settings in all training and evaluation processes. In both iWildCam and RxRx1 datasets, we use the imagenet pretrained ResNet50~\cite{he2016deep} structure for trainings. We compute the average accuracy (avg. acc.) in the two datasets and report Macro F1 for iWildCam. The results are computed over a total of 3 trials in each dataset with different random seeds once the trainings are finished; For Camelyon17 and FMoW, we use the imagenet pretrained DensNet121~\cite{huang2017densely} for the training processes. We report the average accuracy over 10 trials for evaluating Camelyon17, and we use both the metrics of worst-case accuracy and average accuracy to evaluate FMoW, which are computed across 3 trials with varying random seeds.

\begin{table*}[t]
\centering
\caption{Comparison between different invariance constraint and mean value updating schemes in the unseen domain from the PACS~\cite{li2017deeper}, OfficeHome~\cite{venkateswara2017deep}, and DomainNet~\cite{peng2019moment} datasets. Here the ``$Z$'', ``$O$'', and ``$R$'' denotes the feature-invariance, logits-invariance, and the proposed rationale invariance constraints; $\textbf{0}$ is an all-zero tensor, $m$ is the momentum value, when $m=0$, $\overline{R}$ is fixed as the rationale from the pretrained model (\ie Pr.), and when $m=1$, $\overline{R}$ takes the dynamic mean value from the current batch (\ie Dy.). Mt. denotes the proposed momentum updating strategy for obtaining $\overline{R}$. Reported accuracies and standard deviations are computed same as that detailed in Sec.~\ref{sec set}.}
\scalebox{0.98}{
\begin{tabular}{l@{\extracolsep{4pt}}ccc|cccc|cccc@{}}
\toprule 
 \multirow{2}*{Models} & \multicolumn{3}{c|}{Invariance} & \multicolumn{4}{c|}{Setting of $\overline{R}$} & \multicolumn{3}{c}{Test datasets} &\multirow{2}*{Avg.}\\
 \cline{2-4} \cline{5-7} \cline{8-11}
& $Z$ & $O$ & $R$ &Pr. & Dy. & $\textbf{0}$ & Mt. & PACS & OfficeHome & DomainNet\\
\hline \hline
ERM & $-$ & $-$ & $-$ & $-$ & $-$ &$-$ & $-$  &79.8 $\pm$ 0.4 &60.6 $\pm$ 0.2 &35.3 $\pm$0.1 &58.6\\
\hline
W/ fea. & $\checkmark$ & $-$ & $-$ & $-$ & $-$ &$-$ &$\checkmark$ &81.1 $\pm$ 0.5  &62.6 $\pm$ 0.6 &35.8 $\pm$ 0.1 &59.8\\
W/ log. & $-$ & $\checkmark$ & $-$ & $-$ & $-$ &$-$ &$\checkmark$ &81.5 $\pm$ 0.3  &62.2 $\pm$ 0.2 &35.3 $\pm$ 0.3 &59.7\\
\hline
W/ $m=0$  & $-$ & $-$ & $\checkmark$ & $\checkmark$ & $-$ &$-$ &$-$ &81.6 $\pm$ 0.4 &62.5 $\pm$ 0.1  &35.8 $\pm$ 0.1 &60.0\\
W/ $m=1$  & $-$ & $-$ & $\checkmark$ & $-$ & $\checkmark$ &$-$ &$-$ &82.0 $\pm$ 0.3 &62.8 $\pm$ 0.2  &35.0 $\pm$ 0.5 &59.9\\
W/ $\overline{R}=\textbf{0}$ & $-$ & $-$ & $\checkmark$ & $-$ &$-$ &$\checkmark$ &$-$ &81.2 $\pm$ 0.5 &62.9 $\pm$ 0.3  &34.9 $\pm$ 0.6 &59.7\\
\hline
Ours & $-$ & $-$ & $\checkmark$ & $-$ & $-$ &$-$ & $\checkmark$  &82.8 $\pm$ 0.3 &63.3 $\pm$ 0.1 &36.0 $\pm$ 0.2 &60.7 \\
\bottomrule
\end{tabular}}
\label{tab ana}
\vspace{-0.3 cm}
\end{table*}

\noindent\textbf{Experimental results.}
We list the evaluation results in Table~\ref{tab wilds} where the statistics from the compared arts are directly cited from the public leaderboard\footnote{https://wilds.stanford.edu/leaderboard/}. Similar to the observation in~\cite{gulrajani2020search}, we note most of the sophisticated alternatives perform on par with the baseline ERM in general. In particular, there are certain situations in which nearly all methods are outperformed by ERM. This is particularly evident in the FMoW dataset, where ERM achieves higher average accuracy than all other approaches except for our model. 
Differently, our method achieves the best performance in 2 of the 4 evaluated datasets and outperforms ERM in all 4 benchmarks, with particularly impressive results in the Camelyon17 dataset, where improvements of over 10\% were observed. These findings demonstrate the effectiveness of our proposed method. We note our method performs less effectively than~\cite{sun2016deep} in the iWildCam~\cite{beery2021iwildcam} dataset. This is mainly because classes are highly imbalanced in~\cite{beery2021iwildcam}: some classes contain very few samples. This will result the corresponding $\overline{\mathbf{R}}_k$ to not contain information from other samples (\ie $\overline{\mathbf{R}}_k \approx \mathbf{R}_n$), leading to ineffectiveness when computing $\mathcal{L}_{inv}$. We leave the improvements to future works.  

\section{Analysis and Discussion}
This section analyzes the effectiveness of the proposed rationale concept and the adopted momentum updating strategy by conducting ablation studies on the widely-used PACS~\cite{li2017deeper}, OfficeHome~\cite{venkateswara2017deep}, and DomainNet~\cite{peng2019moment} datasets with the rigorous evaluation settings detailed in Sec.~\ref{sec set}. We provide more analysis in our supplementary material, please refer to it for details.

\subsection{Effectiveness of the New Rationale Concept}
\label{sec ana_rational}
We compare our method with the following two variants to demonstrate the effectiveness of the proposed rationale concept. (1) Ours with feature-invariance constraint (\ie W/ fea.), which replaces our rationale-invariance constraint with the existing feature-invariance strategy~\cite{cha2022domain}. In this setting, we reformulate Eq.~\eqref{eq inv} into $\mathcal{L}_{inv} = \frac{1}{N_b}\sum_k \sum_{\{n|y_n=k\}} \Vert \mathbf{z}_n - \overline{\mathbf{z}}_k \Vert^2$, where $\overline{\mathbf{z}}_k$ is also computed in the same momentum manner;
(2) Ours with logit-invariance constraint (\ie W/ log.), which replaces the original design with the logit-invariance constraint~\cite{pezeshki2021gradient} in the overall framework. For this strategy, the invariance constraint in Eq.~\eqref{eq inv} can be rewritten as $\mathcal{L}_{inv} = \frac{1}{N_b}\sum_k \sum_{\{n|y_n=k\}} \Vert \mathbf{o}_n - \overline{\mathbf{o}}_k \Vert^2$, where $\overline{\mathbf{o}}_k$ is the momentum updated mean value of the logits. Other settings for these two variants are kept the same as our original designs to ensure fair comparisons.

We list the evaluation results in 2nd-3rd rows in Table~\ref{tab ana}. We observe that both these two variants can bring certain improvements to the baseline ERM method. This is because both the two invariance constraints can help the model to obtain robust results to a certain extent, thus improving generalization. However, these two strategies have intrinsic limitations. Specifically, when focusing solely on the features, $\mathcal{L}_{inv}$ may amplify the effects of the irreverent feature elements that with large values but correspond to small weights in the decision-making process, thus diminishing the overall classification performance. 
Meanwhile, since $\mathbf{o}$ is the summation of all the element-wise contributions, enforcing $\mathcal{L}_{inv}$ on $\mathbf{o}$ fails to take the varying effect of each contribution value to the final decision into account, which may end up boosting contribution with small values to ensure the corresponding summation equals the mean value. As a result, the model will be encouraged to emphasize the corresponding irreverent features.
Differently, we can control the fine-grained decision-making process by focusing on the rational concept, thus avoiding the aforementioned limitations. This is also the reason why our method can outperform those two variants in all test datasets.

\subsection{Effectiveness of the Momentum Updating}
We obtain the mean rationale value $\overline{\mathbf{R}}$ with a momentum updating strategy. To examine the effectiveness of this scheme, we compare our original design with the following three variants. 
(1) Ours with $m$ set to be 0 (\ie W/ $m=0$). This variant augments the original design by fixing the momentum in Eq.~\eqref{eq momentum} as 0, which can be regarded as replacing $\overline{\mathbf{R}}$ with the rationale from the initial pretrained model, similar to the strategy in~\cite{cha2022domain};
(2) Ours with $m$ set to be 1 (\ie W/ $m=1$). This scheme fixes $m$ as 1, and it can be considered as using the dynamic mean rationale value from the current batch for $\overline{\mathbf{R}}$;
(3) Ours with $\overline{\mathbf{R}}$ fixed as $\textbf{0}$ (\ie W/ $\overline{\mathbf{R}} = \textbf{0}$), where $\textbf{0}$ is an all-zero tensor, similar to that adopted in \cite{pezeshki2021gradient}. Other settings are also the same as the original designs.

\def\swthree{0.27\linewidth}
\renewcommand{\tabcolsep}{-6pt}
\begin{figure*}[t]
\centering
    \begin{tabular}{cccc}
        \includegraphics[width=\swthree]{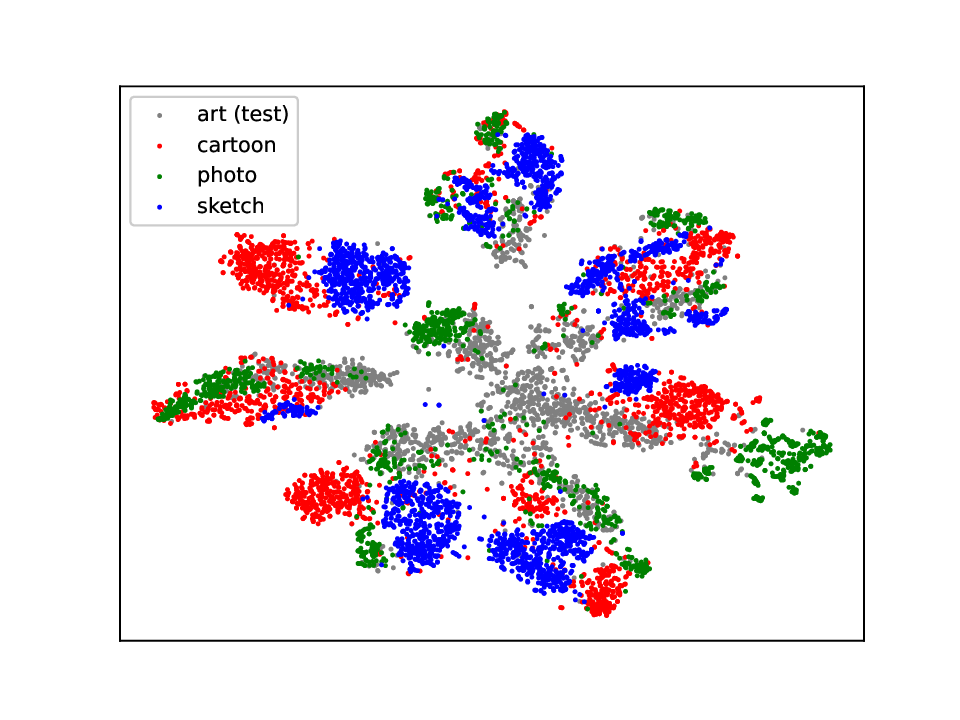}&
        \includegraphics[width=\swthree]{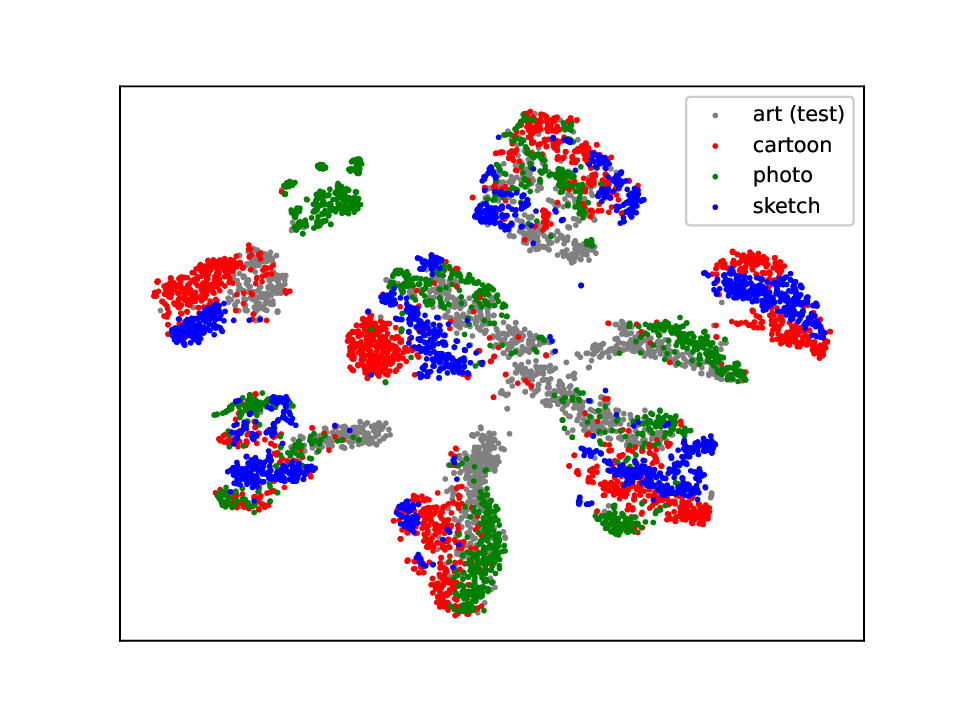}&
        \includegraphics[width=\swthree]{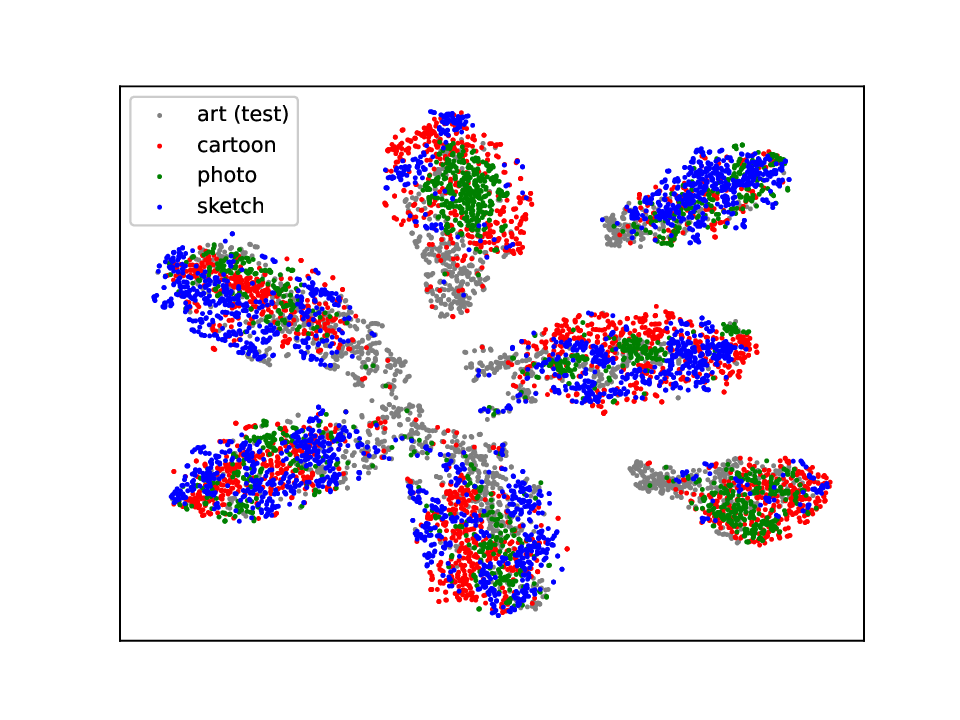}&
        \includegraphics[width=\swthree]{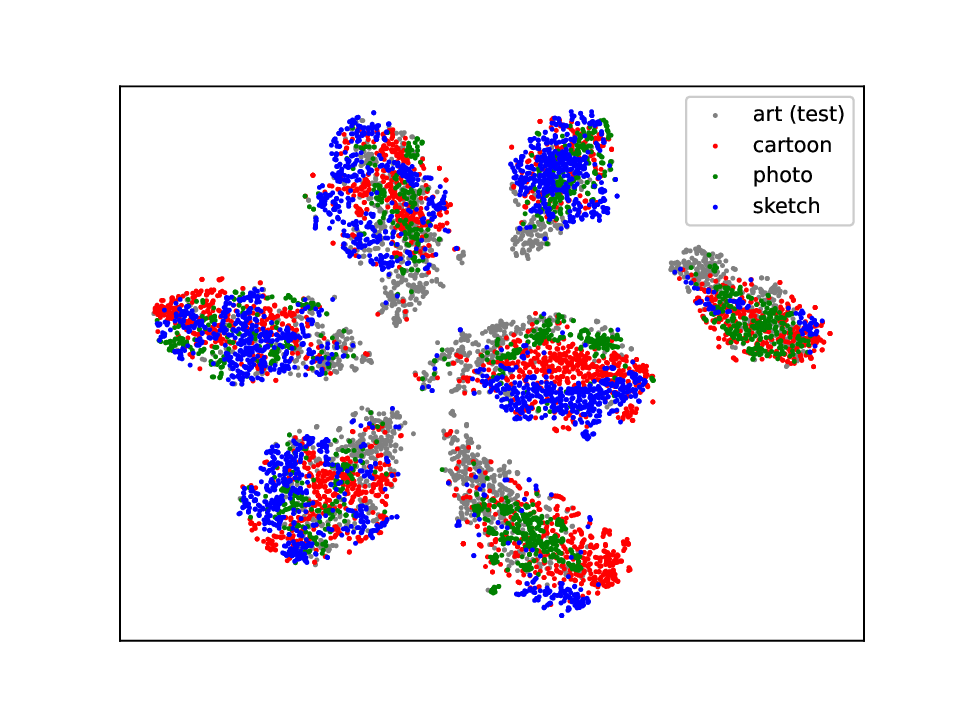} \vspace{-0.4cm}\\
        {\text{(a) ERM}} & {\text{(b) Logit invariance}}& {\text{(c) Feature invariance}} & {\text{(d) Ours}}\\
    \end{tabular}
	\caption{2D t-SNE \cite{van2008visualizing} of the vectorized $\mathbf{R}$ from different invariance regularization. The PACS dataset \cite{li2017deeper} is used with art as the unseen target domain. The seven clusters in the figures denote the corresponding classes. Objects from different categories are well separated in the proposed rationale-based model, yet the corresponding domain information is subtle in each cluster, demonstrating its efficacy in making effective and domain-agnostic decisions.}
	\label{fig tsne}
 \vspace{-0.3 cm}
\end{figure*}

Results regarding these variants are listed in the 4th-6th rows in Table~\ref{tab ana}. We observe that the setting of fixing $\overline{\mathbf{R}}$ as $\textbf{0}$ performs inferior to others, and the performance in each domain is close to the setting of replacing the original design with the logit invariance constraint (\ie W/ log.), this is mainly because this setting uses a uniform value to regularize each contribution value, also neglecting the differences of each individual contribution.
Meanwhile, when $\overline{\mathbf{R}}$ takes the value from the preatrained model (\ie W/ $m=0$), the corresponding performance is slightly better than the other two variants, which may suggest that borrowing rational directly from the pretrained model can help making invariant decisions. We leave such exploration to our future works. 
As for the setting of $\overline{\mathbf{R}}$ using the mean value from the current batch (\ie W/ $m=1$). Due to the limited samples in a batch, it fails to take all samples from the same category into account, thus being outperformed by our momentum updating strategy.

\subsection{Visualizations}
\label{sec tsne}
To better showcase the differences between existing invariance regularization terms, this section presents the 2D t-SNE \cite{van2008visualizing} visualizations of the rationale matrices from different strategies using samples from \cite{li2017deeper}, where the different $\mathbf{R}$ are transformed into their vectorized forms (\ie $\mathbf{R} :\in \mathbb{R}^{DK \times 1}$) before the dimension reduction processes.

Results are plotted in Figure~\ref{fig tsne}. We observe the clusters from the proposed rationale invariance regularization are more clearly separated than that from the other three strategies, demonstrating its effectiveness in making effective decisions when encountering data from an unknown distribution. This observation complies with the results reported in Table~\ref{tab ana}.
Meanwhile, we note that representations from the logit-invariance constraint are also separated by their domain information, similar to that in the baseline ERM method. This is not surprising, as the logit cannot provide fine-grained representations of the decision-making process, thus incapable of obtaining a domain-invariant rationale matrix. 
Note that both the visualizations from our rationale invariance regularization and the classical feature invariance constraint~\cite{ben2006analysis} contain subtle domain information compared to the ERM method, indicating that both the term can enforce the model to make domain-invariant decisions. In order to avoid the potential issue of the feature invariance constraint causing the model to rely heavily on irrelevant features and consequently leading to poor classification results, we thus introduce the new rationale invariance constraint to ensure a robust and effective decision-making process. We provide more visual illustrations in our supplementary material for better comprehending our rational invariance regularization, please refer to it for details.         

\subsection{Limitations and Future Works}
Despite the simplicity and effectiveness of the proposed rationale concept, there remain improvements that can benefit future studies.
First, we note the rationale matrix designed in Eq.~\eqref{eq rationale} is for the general classification task that ends the model by outputting logits. There are certain occasions where it cannot be trivially extended, such as the cases in the regression task where the last layer of the model is a convolution operation. Our future work aims to broaden the application of the rationale concept to a wider range of contexts. 
Second, we implement the invariance constraint by assigning a mean rationale matrix for each class. However, this setting cannot be applied to the situation where the class number is indefinite, such as the Poverty map~\cite{yeh2020using} estimation task in Wilds~\cite{koh2021wilds}. A promising improvement for the proposed framework will be applying it to tasks with continuous labels.   

\section{Conclusion}
This work proposes a simple yet effective method to ease the domain generalization problem. Our method derives from the intuition that a well-generalized model should make robust decisions encountering varying environments. To implement this idea, we introduce the rationale concept, which can be represented as a matrix that collects all the element-wise contributions to the decision-making process for a given sample. To ensure robust outputs, we suggest that the rationale matrices from the same category should remain unchanged, and the idea is fulfilled by enforcing the rationale matrix from a sample to be similar to its corresponding mean value, which is momentum updating during the training process. The overall framework is easy to implement, requiring only a few lines of code upon the baseline. 
Through extensive experiments on existing benchmarks, we demonstrate that the proposed method can consistently improve the baseline and obtain favorable performances against state-of-the-art models, despite its simplicity.

\noindent\textbf{Acknowledgements.} Liang Chen is supported by the China Scholarship Council (CSC Student ID 202008440331).

{\small
\bibliographystyle{ieee_fullname}
\bibliography{dg}
}

\clearpage
\section*{Supplementary Material}
In this supplementary material, we provide,

1. Visualizations of Values from Eq.~(3) in the manuscript in Sec.~\ref{sec visual};

2. Sensitive analysis regarding the hyper-parameters in Sec.~\ref{sec params};

3. Comparison regarding the setting of combing logit and features in Sec.~\ref{sec ana};

4. Evaluations on the DomainBed benchmark using the ResNet50 backbone in Sec.~\ref{sec res50};

\section{Visualizations of Values from Eq.~(3)}
\label{sec visual}
In this section, we plot the changes in the sample-to-center-difference (SCD) values for rationales, features, and logits in Fig.~\ref{fig plot} (a)-(c) in settings of with and without $\mathcal{L}_{inv}$. Our observations are as follows: (1) Using $\mathcal{L}_{inv}$ tends to decrease the three SCD values, which is significant compared to disabling $\mathcal{L}_{inv}$. The results indicate that ERM fails to summarize shared clues to make a robust decision for samples from the same class, explaining why it is less effective in generalizing than ours. (2) When compared to the case of rationales, features, and logits, the SCD values exhibit larger variances throughout iterations, indicating that our $\mathcal{L}_{inv}$ allows for some flexibility, enabling features and logits to deviate from their centers. This observation aligns with our suggestion: the contribution of each feature dimension should be jointly modulated by both the feature itself and its corresponding classifier weight.

 \def\fr{0.32\linewidth}
\begin{figure}[t]
\centering
    \begin{tabular}{ccc}
        \includegraphics[width=\fr]{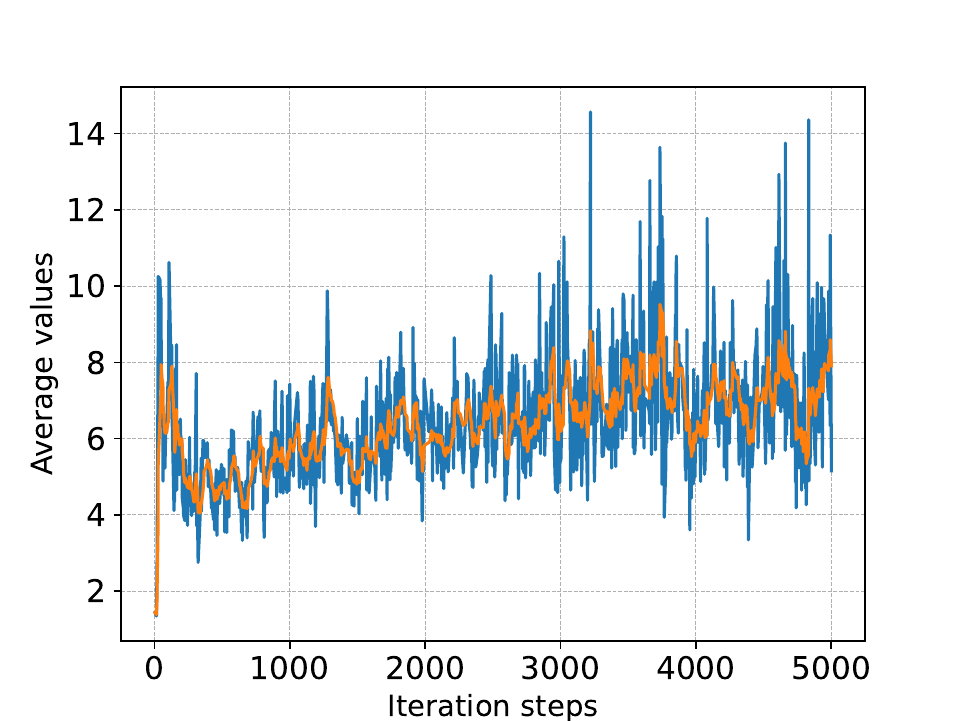} & \hspace{0.5 cm}
        \includegraphics[width=\fr]{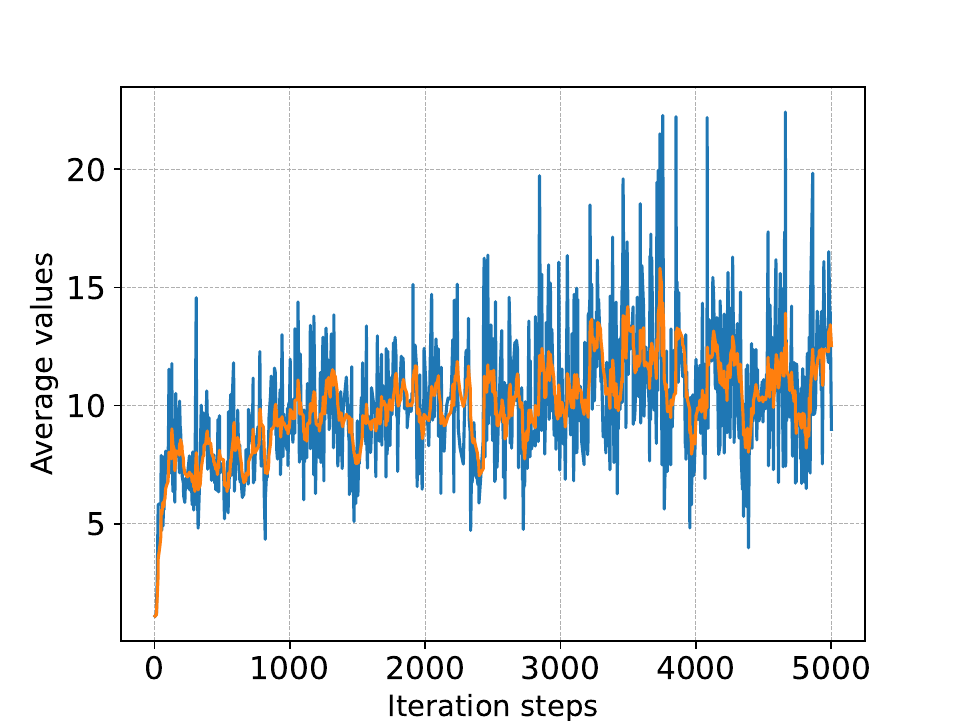}& \hspace{0.5 cm}
        \includegraphics[width=\fr]{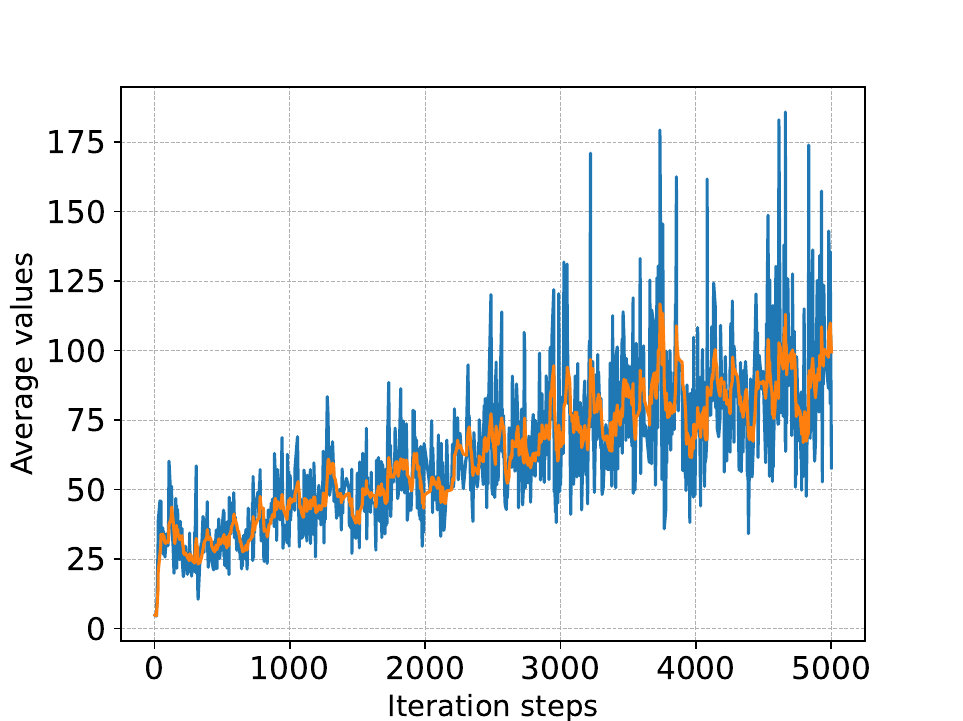}\\
        \includegraphics[width=\fr]{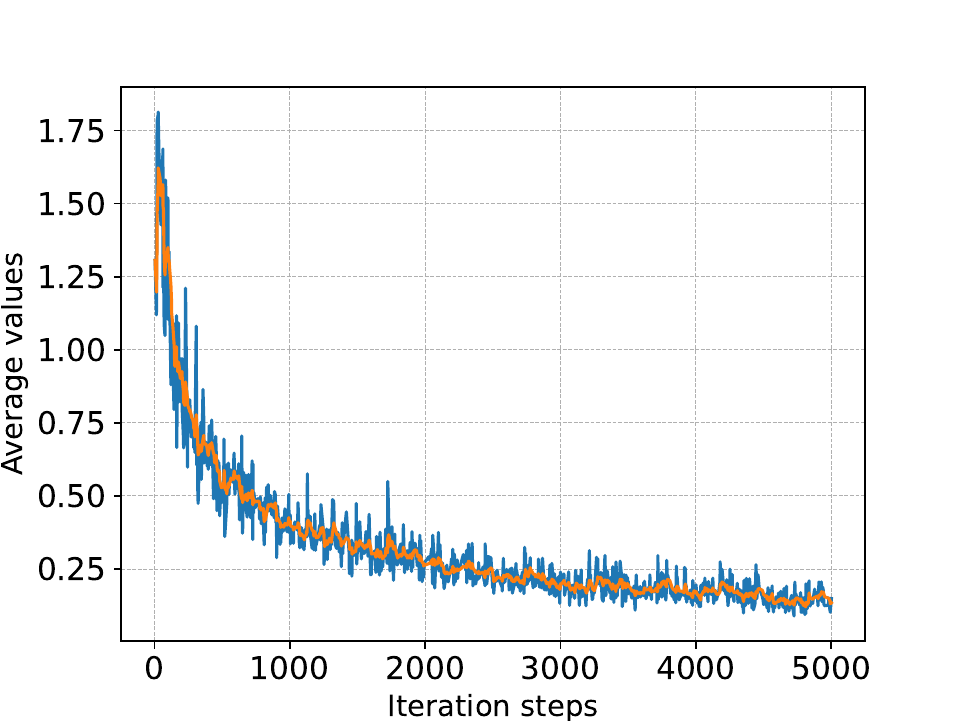}& \hspace{0.5 cm}
        \includegraphics[width=\fr]{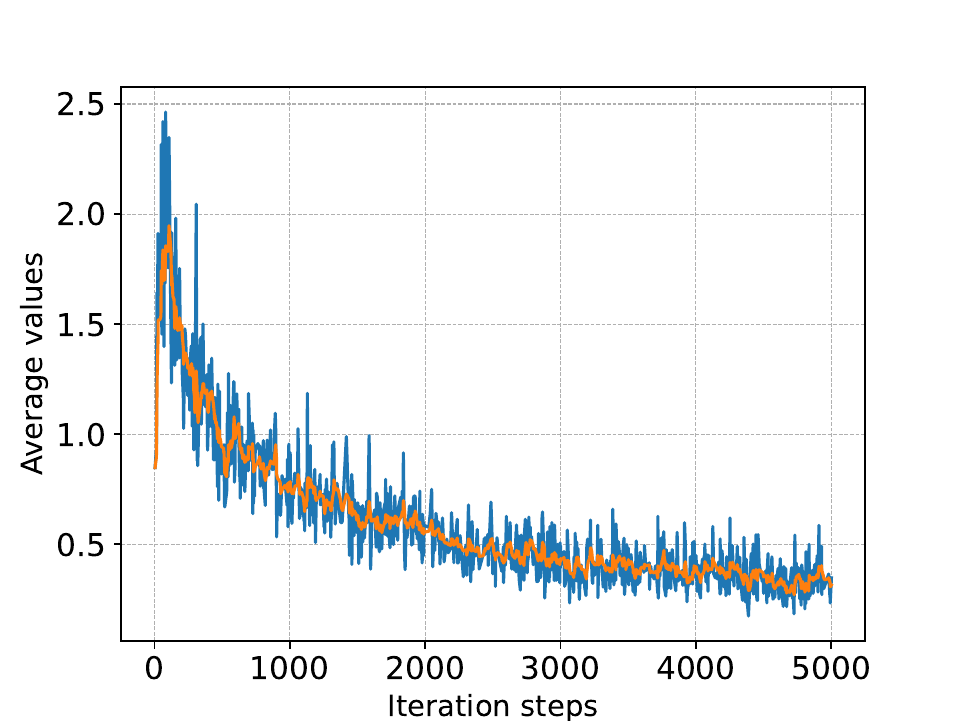}& \hspace{0.5 cm}
        \includegraphics[width=\fr]{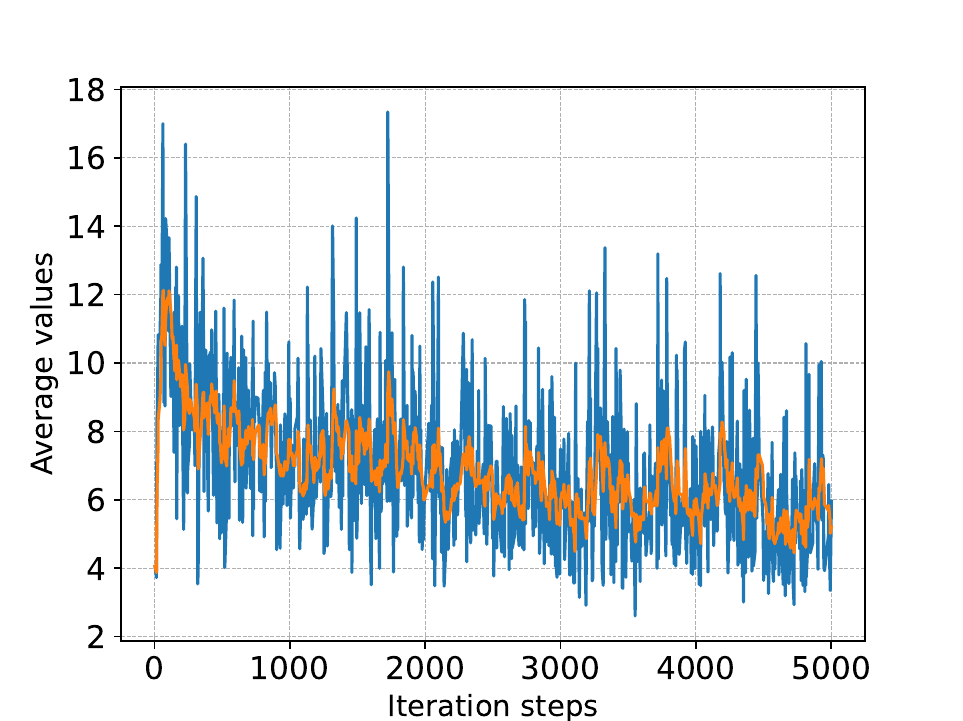}\\
        \scriptsize{\text{(a)}} & \scriptsize{\text{(b)}}& \scriptsize{\text{(c)}} \\
    \end{tabular}
    \caption{Plots of different terms by iterations (\ie Values of Eq. (3), feature, and logit differences in (a)-(c), yellow and blue lines are smoothed and original data). }
	\label{fig plot}
 \end{figure}

\section{Sensitive Analysis Regarding the Hyper-Parameter Settings}
\label{sec params}
Our implementation involves two hyper-parameters: the momentum value $m$ in Eq. (4) and the positive weight $\alpha$ in Eq. (5) in the manuscript. 
This section evaluates our method with different settings of these two hyper-parameters by conducting experiments on the widely-used PACS dataset \cite{li2017deeper} with a ResNet18 backbone \cite{he2016deep} using the same setting illustrated in Sec. 4.1 in the manuscript, similar to that in \cite{chen2023improved}.
Note we fix the value for one hyper-parameter when analyzing another. Results are listed in Table \ref{tab params}. We observe that our method performs consistently well when the hyper-parameter $m$ in the range of $\left[0.0001, 0.1 \right]$ and $\alpha$ in the range of $\left[0.001, 0.1 \right]$. 

\renewcommand{\tabcolsep}{5pt}
\begin{table*}
\centering
\caption{Evaluations regarding different hyper-parameter (\ie $m$ in Eq.~(4) and $\alpha$ in Eq.~(5) from the manuscript) settings. We fix one parameter and tune another when conducting the experiments which are examined in PACS \cite{li2017deeper} with the leave-one-out training-test strategy. The reported accuracies ($\%$) and standard deviations are computed from 60 trials in each target domain.}
\scalebox{1}{
\begin{tabular}{llccccc}
\toprule 
\multicolumn{2}{c}{hyper-parameters} & art & cartoon & photo & sketch &avg \\
\hline \hline
\multirow{6}{*}{$\alpha \in \left[0.001, 0.1 \right]$} & $m=0$ &82.3 $\pm$ 0.1 &73.4 $\pm$ 1.2  &95.0 $\pm$ 0.5 &75.8 $\pm$ 0.8 &81.6 $\pm$ 0.4\\
&$m=0.0001$ &82.3 $\pm$ 0.5 &76.0 $\pm$ 0.5 &94.6 $\pm$ 0.4 &75.9 $\pm$ 1.1 &82.2 $\pm$ 0.4\\
&$m=0.001$ &82.1 $\pm$ 1.4 &75.5 $\pm$ 1.2 &94.9 $\pm$ 0.5 &76.5 $\pm$ 0.3 &82.2 $\pm$ 0.5\\
&$m=0.01$ &82.9 $\pm$ 0.8 &76.2 $\pm$ 1.2 &94.6 $\pm$ 0.6 &75.9 $\pm$ 1.5 &82.4 $\pm$ 5.9\\
&$m=0.1$ &82.2 $\pm$ 0.7 &75.9 $\pm$ 0.9 &95.3 $\pm$ 0.2 &78.1 $\pm$ 0.9 &82.9 $\pm$ 5.9\\
&$m=1$ &81.6 $\pm$ 1.9 &76.2 $\pm$ 0.6  &94.9 $\pm$ 0.4 &75.3 $\pm$ 1.6 &82.0 $\pm$ 0.3\\
\hline
\multirow{5}{*}{$m \in \left[0.0001, 0.1 \right]$} 
&$\alpha=0.0001$ &78.3 $\pm$ 0.6 &74.2 $\pm$ 1.8 &94.0 $\pm$ 0.4 &76.4 $\pm$ 2.4 &80.7 $\pm$ 0.8 \\
&$\alpha=0.001$ &82.3 $\pm$ 0.5 &74.7 $\pm$ 1.4 &93.7 $\pm$ 0.8 &75.5 $\pm$ 0.3 &81.6 $\pm$ 0.4\\
&$\alpha=0.01$ &81.9 $\pm$ 1.0 &75.0 $\pm$ 1.1 &94.9 $\pm$ 0.3 &75.7 $\pm$ 0.4 &81.9 $\pm$ 0.1\\
&$\alpha=0.1$ &82.3 $\pm$ 0.8 &76.2 $\pm$ 0.5 &94.9 $\pm$ 0.4 &76.4 $\pm$ 1.0 &82.4 $\pm$ 0.4\\
&$\alpha=1$ &81.6 $\pm$ 0.7 &74.0 $\pm$ 0.5 &94.5 $\pm$ 0.3 &73.5 $\pm$ 1.4 &80.9 $\pm$ 0.2\\
\bottomrule
\end{tabular}}
\label{tab params}
\end{table*}

\section{Comparisons with the Setting Combining Logit and Feature}
\label{sec ana}
As stated in the manuscript, analyzing the decision-making process from either the perspective of feature or logit has intrinsic limitations. Specifically, since the classifier is not taken into account, the model may emphasize heavily on feature elements that with large values but correspond to small weights in the classifier if only consider the feature. Although logit can ease the issue to a certain extent, it only provides a coarse representation for the decision-making process, thus difficult to ensure robust outputs. 
One may wonder if the combination of feature and logit could avoid the limitation of each other and lead to certain improvements. To answer this question, we conduct further analysis by substituting the rationale invariance constraint with the regularization term that enforces invariance for both the feature and logit (\ie \textbf{W/ fea. \& log.}), which reformulates Eq. (3) into $\mathcal{L}_{inv} = \frac{1}{N_b}\sum_k \sum_{\{n|y_n=k\}} (\Vert \mathbf{z}_n - \overline{\mathbf{z}}_k \Vert^2 + \Vert \mathbf{o}_n - \overline{\mathbf{o}}_k \Vert^2)$, where $\mathbf{z}$, $\mathbf{o}$, $\overline{\mathbf{z}}$ and $\overline{\mathbf{o}}$ are the feature, logit, and their corresponding momentum updated mean values, respectively. We use the same setting as the original design and test the model in the widely-used PACS dataset~\cite{li2017deeper} to evaluate its effectiveness. 

Experimental results are listed in Table~\ref{tab comb}. We note that combining the feature and logit can lead to improvements for both the two invariance constraints (\ie W/ fea. and W/ log.) in almost all target domains. This finding is not surprising since the combined setting considers both the classifier and the feature, thereby mitigating some of the limitations of the two individual settings. However, our rationale invariance regularization still outperforms the combined approach. This is because our rationale concept provides a direct characterization of the decision-making process, encompassing the fine-grained representations of both the features and the weights in the classifier, while the latter can only be coarsely represented in the combined setting.

\begin{table*}[t]
\centering
\caption{Comparison between different invariance constraint and mean value updating schemes in the unseen domain from the PACS benchmark \cite{li2017deeper}. Here the ``$Z$'', ``$O$'', and ``$R$'' denotes the feature-invariance, logits-invariance, and the proposed rationale invariance constraints. The reported accuracies ($\%$) and standard deviations are computed from 60 trials in each target domain.}
\scalebox{1}{
\begin{tabular}{l@{\extracolsep{4pt}}ccc|ccccc@{}}
\toprule 
 \multirow{2}*{Models} & \multicolumn{3}{c|}{Invariance} & \multicolumn{4}{c}{Target domain} &\multirow{2}*{Avg.}\\
 \cline{2-4} \cline{4-8}
& $Z$ & $O$ & $R$ & Art & Cartoon & Photo & Sketch\\
\hline \hline
ERM & $-$ & $-$ & $-$ &78.0 $\pm$ 1.3 &73.4 $\pm$ 0.8 &94.1 $\pm$ 0.4 &73.6 $\pm$ 2.2 &79.8 $\pm$ 0.4\\
\hline
W/ fea. & $\checkmark$ & $-$ &$-$ &82.3 $\pm$ 0.4 &74.3 $\pm$ 1.0  &94.3 $\pm$ 0.4 &73.6 $\pm$ 1.3 &81.1 $\pm$ 0.5\\
W/ log. & $-$ & $\checkmark$ &$-$ &81.9 $\pm$ 0.5 &75.5 $\pm$ 0.5  &94.8 $\pm$ 0.2 &73.9 $\pm$ 1.3 & 81.5$\pm$ 0.3\\
W/ fea \& log. & $\checkmark$ &$\checkmark$ & $-$ &82.2 $\pm$ 1.0 &75.8 $\pm$ 0.6  &95.6 $\pm$ 0.4 &74.7 $\pm$ 0.8  & 82.1 $\pm$ 0.4\\
Ours & $-$ & $-$ & $\checkmark$  &82.4 $\pm$ 1.0 &76.7 $\pm$ 0.6 &95.3 $\pm$ 0.1 &76.7 $\pm$ 0.3 &82.8 $\pm$ 0.3 \\
\bottomrule
\end{tabular}}
\label{tab comb}
\end{table*}

\section{Results in the DomainBed Benchmark with the ResNet50 Backbone}
\label{sec res50}
To comprehensively examine the effectiveness of the proposed method, we also evaluate our method and the baseline ERM and the top-3 arts in Table 1 in the manuscript using the larger ResNet50 backbone \cite{he2016deep}.
Results are listed in Table~\ref{tab res50}, which are directly cited from \cite{kim2021selfreg}. We note that our method surpasses the baseline ERM model in all datasets and leads by 1.4 in average, while the top 3 methods of the compared models (\ie CORAL \cite{sun2016deep}, SagNet \cite{nam2021reducing}, and SelfReg \cite{kim2021selfreg}) lead their ERM model by 1.3, 0.9 and 0.9 in average. These results indicate that our method can consistently improve the baseline model and perform favorably against existing arts when implemented with a larger ResNet50 backbone.

\begin{table*}[t]
\centering
\caption{Average accuracies on the DomainBed \cite{gulrajani2020search} benchmark using the ResNet50 \cite{he2016deep} backbones. Results without $\dagger$ are directly cited from a previous work~\cite{kim2021selfreg}. Improve. denotes the average improvements with respect to the corresponding ERM model.} 
\scalebox{1}{
\begin{tabular}{lC{1.7cm}C{1.7cm}C{1.7cm}C{1.7cm}C{1.7cm}|C{0.9cm}C{1.2cm}}
\toprule 
& PACS & VLCS & OfficeHome & TerraInc & DomainNet & Avg. &Improve.\\
\hline \hline
ERM \cite{vapnik1999nature} &85.5 $\pm$ 0.2 &77.5 $\pm$ 0.4 &66.5 $\pm$ 0.3 &46.1 $\pm$ 1.8 &40.9 $\pm$ 0.1 &63.3 &-\\
IRM \cite{arjovsky2019invariant} &83.5 $\pm$ 0.8 & 78.5 $\pm$ 0.5 &64.3 $\pm$ 2.2 &47.6 $\pm$ 0.8 &33.9 $\pm$ 2.8 &61.2 &-2.1\\
GroupGRO \cite{sagawa2019distributionally} &84.4 $\pm$ 0.8 & 76.7 $\pm$ 0.6 & 66.0 $\pm$ 0.7 & 43.2 $\pm$ 1.1 & 33.3 $\pm$ 0.2 &60.7 &-2.6\\
Mixup \cite{yan2020improve} &84.6 $\pm$ 0.6 & 77.4 $\pm$ 0.6 & 68.1 $\pm$ 0.3 & 47.9 $\pm$ 0.8 & 39.2 $\pm$ 0.1 &63.4 &+0.1\\
MLDG \cite{li2018learning} &84.9 $\pm$ 1.0 & 77.2 $\pm$ 0.4 & 66.8 $\pm$ 0.6 & 47.7 $\pm$ 0.9 & 41.2 $\pm$ 0.1 &63.6 &+0.3\\
CORAL \cite{sun2016deep} &86.2 $\pm$ 0.3 & 78.8 $\pm$ 0.6 & 68.7 $\pm$ 0.3 & 47.6 $\pm$ 1.0 & 41.5 $\pm$ 0.1 &64.6 &+1.3\\
MMD \cite{li2018domain} &84.6 $\pm$ 0.5 & 77.5 $\pm$ 0.9 & 66.3 $\pm$ 0.1 & 42.2 $\pm$ 1.6 & 23.4 $\pm$ 9.5 &58.8 &-4.5\\
DANN \cite{ganin2016domain} &83.6 $\pm$ 0.4 & 78.6 $\pm$ 0.4 & 65.9 $\pm$ 0.6 & 46.7 $\pm$ 0.5 & 38.3 $\pm$ 0.1 &62.6 &-0.7\\
CDANN \cite{li2018deep} &82.6 $\pm$ 0.9 & 77.5 $\pm$ 0.1 & 65.8 $\pm$ 1.3 & 45.8 $\pm$ 1.6 & 38.3 $\pm$ 0.3 &62.0 &-1.3\\
MTL \cite{blanchard2017domain} &84.6 $\pm$ 0.5 & 77.2 $\pm$ 0.4 & 66.4 $\pm$ 0.5 & 45.6 $\pm$ 1.2 & 40.6 $\pm$ 0.1 &62.9 &-0.4\\
SagNet \cite{nam2021reducing} &86.3 $\pm$ 0.2 & 77.8 $\pm$ 0.5 & 68.1 $\pm$ 0.1 & 48.6 $\pm$ 1.0 & 40.3 $\pm$ 0.1 &64.2 &+0.9\\
ARM \cite{zhang2020adaptive} &85.1 $\pm$ 0.4 & 77.6 $\pm$ 0.3 & 64.8 $\pm$ 0.3 & 45.5 $\pm$ 0.3 & 35.5 $\pm$ 0.2 &61.7 &-1.6\\
VREx \cite {krueger2021out} &84.9 $\pm$ 0.6 & 78.3 $\pm$ 0.2 & 66.4 $\pm$ 0.6 & 46.4 $\pm$ 0.6 & 33.6 $\pm$ 2.9 &61.9 &-2.4\\
RSC \cite{huang2020self} &85.2 $\pm$ 0.9 & 77.1 $\pm$ 0.5 & 65.5 $\pm$ 0.9 & 46.6 $\pm$ 1.0 & 38.9 $\pm$ 0.5 &62.7 &-0.6\\
SelfReg \cite{kim2021selfreg} &85.6 $\pm$ 0.4 & 77.8 $\pm$ 0.9 & 67.9 $\pm$ 0.7 & 47.0 $\pm$ 0.3 & 42.8 $\pm$ 0.0 &64.2 &+0.9\\
\bottomrule
$\text{ERM}^\dagger$ \cite{vapnik1999nature} &83.1 $\pm$ 0.9 &77.7 $\pm$ 0.8 &65.8 $\pm$ 0.3 &46.5 $\pm$ 0.9 &40.8 $\pm$ 0.2 &62.8 &-\\
$\text{Fish}^\dagger$ \cite{shi2021gradient} &84.0$\pm$0.3 &78.6$\pm$0.1 &67.9$\pm$0.5 &46.6$\pm$0.4 &40.6$\pm$0.2 &63.5 &+0.7\\
$\text{CORAL}^\dagger$ \cite{sun2016deep} &85.0$\pm$0.4 &77.9$\pm$0.2 &68.8$\pm$0.3 &46.1$\pm$1.2 &41.4$\pm$0.0 &63.9 &+1.1\\
$\text{SD}^\dagger$ \cite{pezeshki2021gradient}  &84.4$\pm$0.2 &77.6$\pm$0.4 &68.9$\pm$0.2 &46.4$\pm$2.0 &42.0$\pm$0.2 &63.9 &+1.1\\
$\text{Ours}^\dagger$ &84.7 $\pm$ 0.2 & 77.8 $\pm$ 0.4 &68.6 $\pm$ 0.2 &47.8 $\pm$ 1.1 &41.9 $\pm$ 0.3 &64.2 &+1.4\\
\bottomrule
\end{tabular}}
\label{tab res50}
\end{table*}

\section{Detailed Results in the DomainBed Benchmark \cite{gulrajani2020search}}
\label{sec detail}
this section presents the average accuracy in each domain from different datasets. As shown in Table~\ref{tab pacs}, ~\ref{tab vlcs}, ~\ref{tab office}, ~\ref{tab terainc}, and ~\ref{tab domainnet}, these results are detailed illustrations of the results in Table 1 in our manuscript. For all the experiments, we use the ``training-domain validate set" as the model selection method. A total of 23 methods are examined for 60 trials in each unseen domain, and all methods are trained with the leave-one-out strategy using the ResNet18~\cite{he2016deep} backbones.

We note the MIRO~\cite{cha2022domain} method performs inferior to other arts when evaluated in the PACS dataset, this is mainly because their approach specifically enforces similarity between intermediate features from the model and that from the pretrained backbone, which can be detrimental to the performance when there is a significant distribution shift between the target data and samples used for pertaining. In this case, the distribution shift is particularly noticeable between data from the `cartoon' and `sketch' domains and real photos in imagenet that are adopted for pretraining. 

\renewcommand{\tabcolsep}{1pt}
\begin{table*}[h]
\centering
\caption{Average accuracies on the PACS \cite{li2017deeper} datasets using the default hyper-parameter settings in DomainBed \cite{gulrajani2020search}.}
\scalebox{1}{
\begin{tabular}{lC{2cm}C{2cm}C{2cm}C{2cm}C{2cm}}
\toprule 
& art & cartoon & photo & sketch & Average\\
\hline \hline
ERM \cite{vapnik1999nature} &78.0 $\pm$ 1.3 &73.4 $\pm$ 0.8 &94.1 $\pm$ 0.4 &73.6 $\pm$ 2.2 &79.8 $\pm$ 0.4 \\
IRM \cite{arjovsky2019invariant} &76.9 $\pm$ 2.6 &75.1 $\pm$ 0.7 &94.3 $\pm$ 0.4 &77.4 $\pm$ 0.4 &80.9 $\pm$ 0.5\\
GroupGRO \cite{sagawa2019distributionally} &77.7 $\pm$ 2.6 &76.4 $\pm$ 0.3 &94.0 $\pm$ 0.3 &74.8 $\pm$ 1.3 &80.7 $\pm$ 0.4 \\
Mixup \cite{yan2020improve} &79.3 $\pm$ 1.1 &74.2 $\pm$ 0.3 &94.9 $\pm$ 0.3 &68.3 $\pm$ 2.7 &79.2 $\pm$ 0.9 \\
MLDG \cite{li2018learning} &78.4 $\pm$ 0.7 &75.1 $\pm$ 0.5 &94.8 $\pm$ 0.4 &76.7 $\pm$ 0.8 &81.3 $\pm$ 0.2 \\
CORAL \cite{sun2016deep} &81.5 $\pm$ 0.5 &75.4 $\pm$ 0.7 &95.2 $\pm$ 0.5 &74.8 $\pm$ 0.4 &81.7 $\pm$ 0.0 \\
MMD \cite{li2018domain} &81.3 $\pm$ 0.6 &75.5 $\pm$ 1.0 &94.0 $\pm$ 0.5 &74.3 $\pm$ 1.5 &81.3 $\pm$ 0.8 \\
DANN \cite{ganin2016domain} &79.0 $\pm$ 0.6 &72.5 $\pm$ 0.7 &94.4 $\pm$ 0.5 &70.8 $\pm$ 3.0 &79.2 $\pm$ 0.3 \\
CDANN \cite{li2018deep} &80.4 $\pm$ 0.8 &73.7 $\pm$ 0.3 &93.1 $\pm$ 0.6 &74.2 $\pm$ 1.7 &80.3 $\pm$ 0.5 \\
MTL \cite{blanchard2017domain} &78.7 $\pm$ 0.6 &73.4 $\pm$ 1.0 &94.1 $\pm$ 0.6 &74.4 $\pm$ 3.0 &80.1 $\pm$ 0.8 \\
SagNet \cite{nam2021reducing} &82.9 $\pm$ 0.4 &73.2 $\pm$ 1.1 &94.6 $\pm$ 0.5 &76.1 $\pm$ 1.8 &81.7 $\pm$ 0.6 \\
ARM \cite{zhang2020adaptive} &79.4 $\pm$ 0.6 &75.0 $\pm$ 0.7 &94.3 $\pm$ 0.6 &73.8 $\pm$ 0.6 &80.6 $\pm$ 0.5 \\
VREx \cite {krueger2021out} &74.4 $\pm$ 0.7 &75.0 $\pm$ 0.4 &93.3 $\pm$ 0.3 &78.1 $\pm$ 0.9 &80.2 $\pm$ 0.5 \\
RSC \cite{huang2020self} &78.5 $\pm$ 1.1 &73.3 $\pm$ 0.9 &93.6 $\pm$ 0.6 &76.5 $\pm$ 1.4 &80.5 $\pm$ 0.2 \\
SelfReg \cite{kim2021selfreg} &82.5 $\pm$ 0.8 &74.4 $\pm$ 1.5 &95.4 $\pm$ 0.5 &74.9 $\pm$ 1.3 &81.8 $\pm$ 0.3 \\
MixStyle \cite{zhou2021domain} &82.6 $\pm$ 1.2 &76.3 $\pm$ 0.4 &94.2 $\pm$ 0.3 &77.5 $\pm$ 1.3 &82.6 $\pm$ 0.4 \\
Fish \cite{shi2021gradient} &80.9 $\pm$ 1.0 &75.9 $\pm$ 0.4 &95.0 $\pm$ 0.4 & 76.2 $\pm$ 1.0 &82.0 $\pm$ 0.3 \\
SD \cite{pezeshki2021gradient} &83.2 $\pm$ 0.6 &74.6 $\pm$ 0.3 &94.6 $\pm$ 0.1 &75.1 $\pm$ 1.6 &81.9 $\pm$ 0.3 \\
CAD \cite{ruan2021optimal} &83.9 $\pm$ 0.8 &74.2 $\pm$ 0.4 &94.6 $\pm$ 0.4 &75.0 $\pm$ 1.2 &81.9 $\pm$ 0.3 \\
CondCAD \cite{ruan2021optimal} &79.7 $\pm$ 1.0 &74.2 $\pm$ 0.9 &94.6 $\pm$ 0.4 &74.8 $\pm$ 1.4 &80.8 $\pm$ 0.5 \\
Fishr \cite{rame2021ishr} &81.2 $\pm$ 0.4 &75.8 $\pm$ 0.8 &94.3 $\pm$ 0.3 &73.8 $\pm$ 0.6 &81.3 $\pm$ 0.3\\
MIRO \cite{cha2022domain} &79.3 $\pm$ 0.6 &68.1 $\pm$ 2.5 &95.5 $\pm$ 0.3 &60.6 $\pm$ 3.1 &75.9 $\pm$ 1.4\\
Ours &82.4 $\pm$ 1.0 &76.7 $\pm$ 0.6 &95.3 $\pm$ 0.1 &76.7 $\pm$ 0.3 &82.8 $\pm$ 0.3 \\
\bottomrule
\end{tabular}}
\label{tab pacs}
\end{table*}

\begin{table*}[h]
\centering
\caption{Average accuracies on the VLCS \cite{fang2013unbiased} datasets using the default hyper-parameter settings in DomainBed \cite{gulrajani2020search}.}
\scalebox{1}{
\begin{tabular}{lC{2cm}C{2cm}C{2cm}C{2cm}C{2cm}}
\toprule 
&Caltech  & LabelMe &Sun & VOC & Average\\
\hline \hline
ERM \cite{vapnik1999nature} &97.7 $\pm$ 0.3 &62.1 $\pm$ 0.9 &70.3 $\pm$ 0.9 &73.2 $\pm$ 0.7 &75.8 $\pm$ 0.2 \\
IRM \cite{arjovsky2019invariant} &96.1 $\pm$ 0.8 &62.5 $\pm$ 0.3 &69.9 $\pm$ 0.7 &72.0 $\pm$ 1.4 &75.1 $\pm$ 0.1\\
GroupGRO \cite{sagawa2019distributionally} &96.7 $\pm$ 0.6 &61.7 $\pm$ 1.5 &70.2 $\pm$ 1.8 &72.9 $\pm$ 0.6 &75.4 $\pm$ 1.0 \\
Mixup \cite{yan2020improve} &95.6 $\pm$ 1.5 &62.7 $\pm$ 0.4 &71.3 $\pm$ 0.3 &75.4 $\pm$ 0.2 &76.2 $\pm$ 0.3 \\
MLDG \cite{li2018learning} &95.8 $\pm$ 0.5 &63.3 $\pm$ 0.8 &68.5 $\pm$ 0.5 &73.1 $\pm$ 0.8 &75.2 $\pm$ 0.3 \\
CORAL \cite{sun2016deep} &96.5 $\pm$ 0.3 &62.8 $\pm$ 0.1 &69.1 $\pm$ 0.6 &73.8 $\pm$ 1.0 &75.5 $\pm$ 0.4 \\
MMD \cite{li2018domain} &96.0 $\pm$ 0.8 &64.3 $\pm$ 0.6 &68.5 $\pm$ 0.6 &70.8 $\pm$ 0.1 &74.9 $\pm$ 0.5 \\
DANN \cite{ganin2016domain} &97.2 $\pm$ 0.1 &63.3 $\pm$ 0.6 &70.2 $\pm$ 0.9 &74.4 $\pm$ 0.2 &76.3 $\pm$ 0.2 \\
CDANN \cite{li2018deep} &95.4 $\pm$ 1.2 &62.6 $\pm$ 0.6 &69.9 $\pm$ 1.3 &76.2 $\pm$ 0.5 &76.0 $\pm$ 0.5 \\
MTL \cite{blanchard2017domain} &94.4 $\pm$ 2.3 &65.0 $\pm$ 0.6 &69.6 $\pm$ 0.6 &71.7 $\pm$ 1.3 &75.2 $\pm$ 0.3 \\
SagNet \cite{nam2021reducing} &94.9 $\pm$ 0.7 &61.9 $\pm$ 0.7 &69.6 $\pm$ 1.3 &75.2 $\pm$ 0.6 &75.4 $\pm$ 0.8 \\
ARM \cite{zhang2020adaptive} &96.9 $\pm$ 0.5 &61.9 $\pm$ 0.4 &71.6 $\pm$ 0.1 &73.3 $\pm$ 0.4 &75.9 $\pm$ 0.3 \\
VREx \cite {krueger2021out} &96.2 $\pm$ 0.0 &62.5 $\pm$ 1.3 &69.3 $\pm$ 0.9 &73.1 $\pm$ 1.2 &75.3 $\pm$ 0.6 \\
RSC \cite{huang2020self} &96.2 $\pm$ 0.0 &63.6 $\pm$ 1.3 &69.8 $\pm$ 1.0 &72.0 $\pm$ 0.4 &75.4 $\pm$ 0.3 \\
SelfReg \cite{kim2021selfreg} &95.8 $\pm$ 0.6 &63.4 $\pm$ 1.1 &71.1 $\pm$ 0.6 &75.3 $\pm$ 0.6 &76.4 $\pm$ 0.7 \\
MixStyle \cite{zhou2021domain} &97.3 $\pm$ 0.3 &61.6 $\pm$ 0.1 &70.4 $\pm$ 0.7 &71.3 $\pm$ 1.9 &75.2 $\pm$ 0.7 \\
Fish \cite{shi2021gradient} &97.4 $\pm$ 0.2 &63.4 $\pm$ 0.1 &71.5 $\pm$ 0.4 &75.2 $\pm$ 0.7 &76.9 $\pm$ 0.2 \\
SD \cite{pezeshki2021gradient} &96.5 $\pm$ 0.4 &62.2 $\pm$ 0.0 &69.7 $\pm$ 0.9 &73.6 $\pm$ 0.4 &75.5 $\pm$ 0.4 \\
CAD \cite{ruan2021optimal} &94.5 $\pm$ 0.9 &63.5 $\pm$ 0.6 &70.4 $\pm$ 1.2 &72.4 $\pm$ 1.3 &75.2 $\pm$ 0.6 \\
CondCAD \cite{ruan2021optimal} &96.5 $\pm$ 0.8 &62.6 $\pm$ 0.4 &69.1 $\pm$ 0.2 &76.0 $\pm$ 0.2 &76.1 $\pm$ 0.3 \\
Fishr \cite{rame2021ishr} &97.2 $\pm$ 0.6 &63.3 $\pm$ 0.7 &70.4 $\pm$ 0.6 &74.0 $\pm$ 0.8 &76.2 $\pm$ 0.3\\
MIRO \cite{cha2022domain} &97.5 $\pm$ 0.2 &62.0 $\pm$ 0.5 &71.3 $\pm$ 1.0 &74.8 $\pm$ 0.6 &76.4 $\pm$ 0.4\\
Ours &96.7 $\pm$ 0.5 &63.2 $\pm$ 1.0 &70.3 $\pm$ 0.8 &73.4 $\pm$ 0.3  &75.9 $\pm$ 0.3 \\
\bottomrule
\end{tabular}}
\label{tab vlcs}
\end{table*}

\begin{table*}
\centering
\caption{Average accuracies on the OfficeHome \cite{venkateswara2017deep} datasets using the default hyper-parameter settings in DomainBed \cite{gulrajani2020search}.}
\scalebox{1}{
\begin{tabular}{lC{2cm}C{2cm}C{2cm}C{2cm}C{2cm}}
\toprule 
&art  & clipart &product & real & Average\\
\hline \hline
ERM \cite{vapnik1999nature} &52.2 $\pm$ 0.2 &48.7 $\pm$ 0.5 &69.9 $\pm$ 0.5 &71.7 $\pm$ 0.5 &60.6 $\pm$ 0.2 \\
IRM \cite{arjovsky2019invariant} &49.7 $\pm$ 0.2 &46.8 $\pm$ 0.5 &67.5 $\pm$ 0.4 &68.1 $\pm$ 0.6 &58.0 $\pm$ 0.1\\
GroupGRO \cite{sagawa2019distributionally} &52.6 $\pm$ 1.1 &48.2 $\pm$ 0.9 &69.9 $\pm$ 0.4 &71.5 $\pm$ 0.8 &60.6 $\pm$ 0.3 \\
Mixup \cite{yan2020improve} &54.0 $\pm$ 0.7 &49.3 $\pm$ 0.7 &70.7 $\pm$ 0.7 &72.6 $\pm$ 0.3 &61.7 $\pm$ 0.5 \\
MLDG \cite{li2018learning} &53.1 $\pm$ 0.3 &48.4 $\pm$ 0.3 &70.5 $\pm$ 0.7 &71.7 $\pm$ 0.4 &60.9 $\pm$ 0.2 \\
CORAL \cite{sun2016deep} &55.1 $\pm$ 0.7 &49.7 $\pm$ 0.9 &71.8 $\pm$ 0.2 &73.1 $\pm$ 0.5 &62.4 $\pm$ 0.4 \\
MMD \cite{li2018domain} &50.9 $\pm$ 1.0 &48.7 $\pm$ 0.3 &69.3 $\pm$ 0.7 &70.7 $\pm$ 1.3 &59.9 $\pm$ 0.4 \\
DANN \cite{ganin2016domain} &51.8 $\pm$ 0.5 &47.1 $\pm$ 0.1 &69.1 $\pm$ 0.7 &70.2 $\pm$ 0.7 &59.5 $\pm$ 0.5 \\
CDANN \cite{li2018deep} &51.4 $\pm$ 0.5 &46.9 $\pm$ 0.6 &68.4 $\pm$ 0.5 &70.4 $\pm$ 0.4 &59.3 $\pm$ 0.4 \\
MTL \cite{blanchard2017domain} &51.6 $\pm$ 1.5 &47.7 $\pm$ 0.5 &69.1 $\pm$ 0.3 &71.0 $\pm$ 0.6 &59.9 $\pm$ 0.5 \\
SagNet \cite{nam2021reducing} &55.3 $\pm$ 0.4 &49.6 $\pm$ 0.2 &72.1 $\pm$ 0.4 &73.2 $\pm$ 0.4 &62.5 $\pm$ 0.3 \\
ARM \cite{zhang2020adaptive} &51.3 $\pm$ 0.9 &48.5 $\pm$ 0.4 &68.0 $\pm$ 0.3 &70.6 $\pm$ 0.1 &59.6 $\pm$ 0.3 \\
VREx \cite {krueger2021out} &51.1 $\pm$ 0.3 &47.4 $\pm$ 0.6 &69.0 $\pm$ 0.4 &70.5 $\pm$ 0.4 &59.5 $\pm$ 0.1 \\
RSC \cite{huang2020self} &49.0 $\pm$ 0.1 &46.2 $\pm$ 1.5 &67.8 $\pm$ 0.7 &70.6 $\pm$ 0.3 &58.4 $\pm$ 0.6 \\
SelfReg \cite{kim2021selfreg} &55.1 $\pm$ 0.8 &49.2 $\pm$ 0.6 &72.2 $\pm$ 0.3 &73.0 $\pm$ 0.3 &62.4 $\pm$ 0.1 \\
MixStyle \cite{zhou2021domain} &50.8 $\pm$ 0.6 &51.4 $\pm$ 1.1 &67.6 $\pm$ 1.3 &68.8 $\pm$ 0.5 &59.6 $\pm$ 0.8 \\
Fish \cite{shi2021gradient} &54.6 $\pm$ 1.0 &49.6 $\pm$ 1.0 &71.3 $\pm$ 0.6 &72.4 $\pm$ 0.2 &62.0 $\pm$ 0.6 \\
SD \cite{pezeshki2021gradient} &55.0 $\pm$ 0.4 &51.3 $\pm$ 0.5 &72.5 $\pm$ 0.2 &72.7 $\pm$ 0.3 &62.9 $\pm$ 0.2 \\
CAD \cite{ruan2021optimal} &52.1 $\pm$ 0.6 &48.3 $\pm$ 0.5 &69.7 $\pm$ 0.3 &71.9 $\pm$ 0.4 &60.5 $\pm$ 0.3 \\
CondCAD \cite{ruan2021optimal} &53.3 $\pm$ 0.6 &48.4 $\pm$ 0.2 &69.8 $\pm$ 0.9 &72.6 $\pm$ 0.1 &61.0 $\pm$ 0.4 \\
Fishr \cite{rame2021ishr} &52.6 $\pm$ 0.9 &48.6 $\pm$ 0.3 &69.9 $\pm$ 0.6 &72.4 $\pm$ 0.4 &60.9 $\pm$ 0.3 \\
MIRO \cite{cha2022domain} &57.4 $\pm$ 0.9 &49.5 $\pm$ 0.3 &74.0 $\pm$ 0.1 &75.6 $\pm$ 0.2 &64.1 $\pm$ 0.4 \\
Ours &56.6 $\pm$ 0.7 &50.3 $\pm$ 0.6 &72.5 $\pm$ 0.0 &73.8 $\pm$ 0.3 &63.3 $\pm$ 0.1 \\
\bottomrule
\end{tabular}}
\label{tab office}
\end{table*}

\newpage
\begin{table*}
\centering
\caption{Average accuracies on the TerraInc \cite{beery2018recognition} datasets using the default hyper-parameter settings in DomainBed \cite{gulrajani2020search}.}
\scalebox{1}{
\begin{tabular}{lC{2cm}C{2cm}C{2cm}C{2cm}C{2cm}}
\toprule 
&L100  & L38 &L43 & L46 & Average\\
\hline \hline
ERM \cite{vapnik1999nature} &42.1 $\pm$ 2.5 &30.1 $\pm$ 1.2 &48.9 $\pm$ 0.6 &34.0 $\pm$ 1.1 &38.8 $\pm$ 1.0 \\
IRM \cite{arjovsky2019invariant} &41.8 $\pm$ 1.8 &29.0 $\pm$ 3.6 &49.6 $\pm$ 2.1 &33.1 $\pm$ 1.5 &38.4 $\pm$ 0.9\\
GroupGRO \cite{sagawa2019distributionally} &45.3 $\pm$ 4.6 &36.1 $\pm$ 4.4 &51.0 $\pm$ 0.8 &33.7 $\pm$ 0.9 &41.5 $\pm$ 2.0 \\
Mixup \cite{yan2020improve} &49.4 $\pm$ 2.0 &35.9 $\pm$ 1.8 &53.0 $\pm$ 0.7          &30.0 $\pm$ 0.9 &42.1 $\pm$ 0.7 \\
MLDG \cite{li2018learning} &39.6 $\pm$ 2.3 &33.2 $\pm$ 2.7 &52.4 $\pm$ 0.5          &35.1 $\pm$ 1.5 &40.1 $\pm$ 0.9 \\
CORAL \cite{sun2016deep} &46.7 $\pm$ 3.2 &36.9 $\pm$ 4.3 &49.5 $\pm$ 1.9          &32.5 $\pm$ 0.7 &41.4 $\pm$ 1.8 \\
MMD \cite{li2018domain} &49.1 $\pm$ 1.2 &36.4 $\pm$ 4.8 &50.4 $\pm$ 2.1          &32.3 $\pm$ 1.5 &42.0 $\pm$ 1.0 \\
DANN \cite{ganin2016domain} &44.3 $\pm$ 3.6 &28.0 $\pm$ 1.5 &47.9 $\pm$ 1.0          &31.3 $\pm$ 0.6 &37.9 $\pm$ 0.9 \\
CDANN \cite{li2018deep} &36.9 $\pm$ 6.4 &32.7 $\pm$ 6.2 &51.1 $\pm$ 1.3          &33.5 $\pm$ 0.5 &38.6 $\pm$ 2.3 \\
MTL \cite{blanchard2017domain} &45.2 $\pm$ 2.6 &31.0 $\pm$ 1.6 &50.6 $\pm$ 1.1          &34.9 $\pm$ 0.4 &40.4 $\pm$ 1.0 \\
SagNet \cite{nam2021reducing} &36.3 $\pm$ 4.7 &40.3 $\pm$ 2.0 &52.5 $\pm$ 0.6          &33.3 $\pm$ 1.3 &40.6 $\pm$ 1.5 \\
ARM \cite{zhang2020adaptive} &41.5 $\pm$ 4.5 &27.7 $\pm$ 2.4 &50.9 $\pm$ 1.0          &29.6 $\pm$ 1.5 &37.4 $\pm$ 1.9 \\
VREx \cite {krueger2021out} &48.0 $\pm$ 1.7 &41.1 $\pm$ 1.5 &51.8 $\pm$ 1.5          &32.0 $\pm$ 1.2 &43.2 $\pm$ 0.3 \\
RSC \cite{huang2020self} &42.8 $\pm$ 2.4 &32.2 $\pm$ 3.8 &49.6 $\pm$ 0.9          &32.9 $\pm$ 1.2 &39.4 $\pm$ 1.3 \\
SelfReg \cite{kim2021selfreg} &46.1 $\pm$ 1.5 &34.5 $\pm$ 1.6 &49.8 $\pm$ 0.3          &34.7 $\pm$ 1.5 &41.3 $\pm$ 0.3 \\
MixStyle \cite{zhou2021domain} &50.6 $\pm$ 1.9 &28.0 $\pm$ 4.5 &52.1 $\pm$ 0.7          &33.0 $\pm$ 0.2 &40.9 $\pm$ 1.1 \\
Fish \cite{shi2021gradient} &46.3 $\pm$ 3.0 &29.0 $\pm$ 1.1 &52.7 $\pm$ 1.2          &32.8 $\pm$ 1.0 &40.2 $\pm$ 0.6 \\
SD \cite{pezeshki2021gradient} &45.5 $\pm$ 1.9 &33.2 $\pm$ 3.1 &52.9 $\pm$ 0.7          &36.4 $\pm$ 0.8 &42.0 $\pm$ 1.0 \\
CAD \cite{ruan2021optimal} &43.1 $\pm$ 2.6 &31.1 $\pm$ 1.9 &53.1 $\pm$ 1.6          &34.7 $\pm$ 1.3 &40.5 $\pm$ 0.4 \\
CondCAD \cite{ruan2021optimal} &44.4 $\pm$ 2.9 &32.9 $\pm$ 2.5 &50.5 $\pm$ 1.3          &30.8 $\pm$ 0.5 &39.7 $\pm$ 0.4 \\
Fishr \cite{rame2021ishr} &49.9 $\pm$ 3.3 &36.6 $\pm$ 0.9 &49.8 $\pm$ 0.2 &34.2 $\pm$ 1.3 &42.6 $\pm$ 1.0 \\
MIRO \cite{cha2022domain} &46.0 $\pm$ 0.7 &34.4 $\pm$ 0.4 &51.2 $\pm$ 1.0 &33.6 $\pm$ 0.9 &41.3 $\pm$ 0.2 \\
Ours &46.2 $\pm$ 4.0 &39.7 $\pm$ 2.4 &53.0 $\pm$ 0.6 &36.0 $\pm$ 0.3 &43.7 $\pm$ 0.5 \\
\bottomrule
\end{tabular}}
\label{tab terainc}
\end{table*}

\begin{table*}
\centering
\caption{Average accuracies on the DomainNet \cite{peng2019moment} datasets using the default hyper-parameter settings in DomainBed  \cite{gulrajani2020search}.}
\scalebox{1}{
\begin{tabular}{lC{2cm}C{2cm}C{2cm}C{2cm}C{2cm}C{2cm}C{2 cm}}
\toprule 
&clip  & info &paint & quick &real &sketch & Average\\
\hline \hline
ERM \cite{vapnik1999nature} &50.4 $\pm$ 0.2 &14.0 $\pm$ 0.2 &40.3 $\pm$ 0.5          &11.7 $\pm$ 0.2 &52.0 $\pm$ 0.2 &43.2 $\pm$ 0.3 &35.3 $\pm$ 0.1 \\
IRM \cite{arjovsky2019invariant} &43.2 $\pm$ 0.9 &12.6 $\pm$ 0.3 &35.0 $\pm$ 1.4          &9.9 $\pm$ 0.4 &43.4 $\pm$ 3.0 &38.4 $\pm$ 0.4 &30.4 $\pm$ 1.0\\
GroupGRO \cite{sagawa2019distributionally} &38.2 $\pm$ 0.5 &13.0 $\pm$ 0.3          &28.7 $\pm$ 0.3 &8.2 $\pm$ 0.1 &43.4 $\pm$ 0.5 &33.7 $\pm$ 0.0 &27.5 $\pm$ 0.1 \\
Mixup \cite{yan2020improve} &48.9 $\pm$ 0.3 &13.6 $\pm$ 0.3 &39.5 $\pm$ 0.5          &10.9 $\pm$ 0.4 &49.9 $\pm$ 0.2 &41.2 $\pm$ 0.2 &34.0 $\pm$ 0.0 \\
MLDG \cite{li2018learning} &51.1 $\pm$ 0.3 &14.1 $\pm$ 0.3 &40.7 $\pm$ 0.3          &11.7 $\pm$ 0.1 &52.3 $\pm$ 0.3 &42.7 $\pm$ 0.2 &35.4 $\pm$ 0.0 \\
CORAL \cite{sun2016deep} &51.2 $\pm$ 0.2 &15.4 $\pm$ 0.2 &42.0 $\pm$ 0.2 &12.7 $\pm$ 0.1 &52.0 $\pm$ 0.3 &43.4 $\pm$ 0.0 &36.1 $\pm$ 0.2 \\
MMD \cite{li2018domain} &16.6 $\pm$ 13.3 &0.3 $\pm$ 0.0 &12.8 $\pm$ 10.4 &0.3 $\pm$ 0.0 &17.1 $\pm$ 13.7 &0.4 $\pm$ 0.0 &7.9 $\pm$ 6.2 \\
DANN \cite{ganin2016domain}  &45.0 $\pm$ 0.2 &12.8 $\pm$ 0.2 &36.0 $\pm$ 0.2          &10.4 $\pm$ 0.3 &46.7 $\pm$ 0.3 &38.0 $\pm$ 0.3 &31.5 $\pm$ 0.1\\
CDANN \cite{li2018deep} &45.3 $\pm$ 0.2 &12.6 $\pm$ 0.2 &36.6 $\pm$ 0.2          &10.3 $\pm$ 0.4 &47.5 $\pm$ 0.1 &38.9 $\pm$ 0.4 &31.8 $\pm$ 0.2 \\
MTL \cite{blanchard2017domain} &50.6 $\pm$ 0.2 &14.0 $\pm$ 0.4 &39.6 $\pm$ 0.3          &12.0 $\pm$ 0.3 &52.1 $\pm$ 0.1 &41.5 $\pm$ 0.0 &35.0 $\pm$ 0.0 \\
SagNet \cite{nam2021reducing} &51.0 $\pm$ 0.1 &14.6 $\pm$ 0.1 &40.2 $\pm$ 0.2          &12.1 $\pm$ 0.2 &51.5 $\pm$ 0.3 &42.4 $\pm$ 0.1 &35.3 $\pm$ 0.1 \\
ARM \cite{zhang2020adaptive} &43.0 $\pm$ 0.2 &11.7 $\pm$ 0.2 &34.6 $\pm$ 0.1          &9.8 $\pm$ 0.4 &43.2 $\pm$ 0.3 &37.0 $\pm$ 0.3 &29.9 $\pm$ 0.1 \\
VREx \cite {krueger2021out} &39.2 $\pm$ 1.6 &11.9 $\pm$ 0.4 &31.2 $\pm$ 1.3          &10.2 $\pm$ 0.4 &41.5 $\pm$ 1.8 &34.8 $\pm$ 0.8 &28.1 $\pm$ 1.0 \\
RSC \cite{huang2020self} &39.5 $\pm$ 3.7 &11.4 $\pm$ 0.8 &30.5 $\pm$ 3.1          &10.2 $\pm$ 0.8 &41.0 $\pm$ 1.4 &34.7 $\pm$ 2.6 &27.9 $\pm$ 2.0 \\
SelfReg \cite{kim2021selfreg} &47.9 $\pm$ 0.3 &15.1 $\pm$ 0.3 &41.2 $\pm$ 0.2          &11.7 $\pm$ 0.3 &48.8 $\pm$ 0.0 &43.8 $\pm$ 0.3 &34.7 $\pm$ 0.2 \\
MixStyle \cite{zhou2021domain} &49.1 $\pm$ 0.4 &13.4 $\pm$ 0.0 &39.3 $\pm$ 0.0 &11.4 $\pm$ 0.4 &47.7 $\pm$ 0.3 &42.7 $\pm$ 0.1 &33.9 $\pm$ 0.1\\
Fish \cite{shi2021gradient} &51.5 $\pm$ 0.3 &14.5 $\pm$ 0.2 &40.4 $\pm$ 0.3          &11.7 $\pm$ 0.5 &52.6 $\pm$ 0.2 &42.1 $\pm$ 0.1 &35.5 $\pm$ 0.0 \\
SD \cite{pezeshki2021gradient} &51.3 $\pm$ 0.3 &15.5 $\pm$ 0.1 &41.5 $\pm$ 0.3          &12.6 $\pm$ 0.2 &52.9 $\pm$ 0.2 &44.0 $\pm$ 0.4 &36.3 $\pm$ 0.2 \\
CAD \cite{ruan2021optimal} &45.4 $\pm$ 1.0 &12.1 $\pm$ 0.5 &34.9 $\pm$ 1.1          &10.2 $\pm$ 0.6 &45.1 $\pm$ 1.6 &38.5 $\pm$ 0.6 &31.0 $\pm$ 0.8 \\
CondCAD \cite{ruan2021optimal} &46.1 $\pm$ 1.0 &13.3 $\pm$ 0.4 &36.1 $\pm$ 1.4          &10.7 $\pm$ 0.2 &46.8 $\pm$ 1.3 &38.7 $\pm$ 0.7 &31.9 $\pm$ 0.7 \\
Fishr \cite{rame2021ishr} &47.8 $\pm$ 0.7 &14.6 $\pm$ 0.2 &40.0 $\pm$ 0.3 &11.9 $\pm$ 0.2 &49.2 $\pm$ 0.7 &41.7 $\pm$ 0.1 &34.2 $\pm$ 0.3 \\
MIRO \cite{cha2022domain} &50.9 $\pm$ 0.2 &15.6 $\pm$ 0.1 &41.9 $\pm$ 0.4 &10.4 $\pm$ 0.1 &55.1 $\pm$ 0.1 &42.5 $\pm$ 0.3 &36.1 $\pm$ 0.1 \\
Ours &50.2 $\pm$ 0.3 &15.9 $\pm$ 0.1 &42.0 $\pm$ 0.5 &12.6 $\pm$ 0.2 &51.3 $\pm$ 0.1 &43.8 $\pm$ 0.3 &36.0 $\pm$ 0.2 \\
\bottomrule
\end{tabular}}
\label{tab domainnet}
\end{table*}

\end{document}